\newif\ifdraft
\draftfalse

\newif\ifanonymous
\anonymousfalse

\PassOptionsToPackage{table}{xcolor}

\documentclass[lettersize,journal]{IEEEtran}

\usepackage{xcolor}  
\usepackage{cite}
\usepackage{amsmath,amssymb,amsfonts}
\usepackage{bbm}
\usepackage{graphicx}

\usepackage{hyperref}
\usepackage{caption,subcaption}
\usepackage{multirow}
\usepackage{cleveref}
\usepackage{comment}
\usepackage{hhline}
\usepackage{caption,subcaption}
\usepackage{siunitx}
\usepackage{booktabs}
\usepackage{csvsimple}

\usepackage{datatool}
\usepackage{array}
\usepackage{placeins}
\usepackage{float}
\usepackage{geometry}       
\usepackage{pifont} 
\usepackage{makecell}

\definecolor{improvement}{RGB}{144,238,144} 
\definecolor{degradation}{RGB}{255,182,193} 
\definecolor{nochange}{RGB}{255,255,255}

\newcommand{\heading}[1]{\noindent\textbf{#1}}

\newcommand{\diffcellsplitside}[4]{%
    \makecell[l]{%
        \colorbox{#1}{\parbox{0.7cm}{\centering \scriptsize #2}} \hspace{0.2cm}%
        \colorbox{#3}{\parbox{0.7cm}{\centering \scriptsize #4}}%
    }
}
\usepackage[normalem]{ulem} 
\newcommand{\stkout}[1]{\ifmmode\text{\sout{\ensuremath{#1}}}\else\sout{#1}\fi}

\newcommand{\citex}[1]{\mbox{\cite{#1}}}

\ifdraft

\newcommand{\added}[1]{\textcolor{blue}{#1}}
\newcommand{\deleted}[1]{\textcolor{red}{\stkout{#1}}}
\newcommand{\replaced}[2]{\textcolor{blue}{#1} \textcolor{red}{\stkout{#2}}}
\newcommand{\deletedfloat}[1]{}
\newcommand{\commented}[1]{\textcolor{blue}{#1}}
\else

\newcommand{\added}[1]{#1}
\newcommand{\deleted}[1]{}
\newcommand{\replaced}[2]{#1}
\newcommand{\deletedfloat}[1]{}
\newcommand{\commented}[1]{}
\fi

\newcommand\citep[1]{\cite{#1}}
\newcommand\citet[1]{\cite{#1}}

\def\BibTeX{{\rm B\kern-.05em{\sc i\kern-.025em b}\kern-.08em
    T\kern-.1667em\lower.7ex\hbox{E}\kern-.125emX}}
\begin{document}

\title{Generalizable deep learning for photoplethysmography-based blood pressure estimation– A Benchmarking Study}
\ifanonymous
\author{Anonymous}
\else
\author{Mohammad Moulaeifard, Peter H. Charlton, and Nils Strodthoff
\thanks{Mohammad Moulaeifard and Nils Strodthoff are with Carl von Ossietzky Universität Oldenburg, Oldenburg, Germany. (email: mohammad.moulaeifard@uol.de, nils.strodthoff@uol.de). Peter H Charlton is with the University of Cambridge, Cambridge, UK. Corresponding author: NS.}}
\fi

\bstctlcite{BSTcontrol}

\maketitle

\begin{abstract}
Photoplethysmography (PPG)-based blood pressure (BP) estimation represents a promising alternative to cuff-based BP measurements. Recently, an increasing number of deep learning models have been proposed to infer BP from the raw PPG waveform. However, these models have been predominantly evaluated on in-distribution (ID) test sets, which immediately raises the question of the generalizability of these models to external datasets. To investigate this question, we trained five deep learning models on the recently released PulseDB dataset, provided ID benchmarking results on this dataset, and then assessed \added{their} out-of-distribution (OOD) performance on several external datasets. The best model (XResNet1d101) achieved ID MAEs of 9.0 and 5.8 mmHg for systolic and diastolic BP, respectively, on PulseDB with subject-specific calibration, and 13.9 and 8.5 mmHg\added{,} respectively\added{,} without calibration. \added{The} equivalent MAEs on external test datasets without calibration ranged from 10.0 to 18.6 mmHg (SBP) and 5.9 to 10.3 mmHg (DBP).  Our results indicate that \deleted{the} performance is strongly influenced by the differences in BP distributions between datasets. We investigated a simple way of improving performance through sample-based domain adaptation and put forward recommendations for training models with good generalization properties. With this work, we hope to educate more researchers \replaced{about}{for} the importance and challenges of OOD generalization. 
 
\end{abstract}


\section{Introduction}
\label{sec:I}

\IEEEPARstart{P}PG devices \replaced{offer}{represent} a promising approach to monitor vital parameters such as blood pressure, heart rate, and respiratory rate. Their non-invasive nature and cost-effectiveness in comparison to alternative approaches make\deleted{s} them popular for both medical and personal health monitoring purposes \cite {charlton20232023, castaneda2018review, gonzalez2023benchmark}. One of the most widely considered prediction problems based on PPG data is blood pressure (BP) estimation.

Traditionally, BP estimation from PPG signals has involved analyzing features in the PPG data and connecting them to BP measurements using different methods, e.g., pulse wave analysis \cite{o2001pulse}. However, the traditional methods possess certain limitations in real-world applications, e.g.,  variations in physiology \cite {allen2007photoplethysmography} and the requirement for calibration \cite{Wan2020,gonzalez2023benchmark}. 

In light of the limitations of traditional feature-based approaches, there has been a gradual shift of interest on the part of researchers towards machine learning (ML) and deep learning (DL) techniques for predicting BP using PPG signals. These methods autonomously extract features from PPG signals \cite{martinez2022data} and typically \replaced{achieve}{reach} higher accuracy in data-rich environments \cite{chang2021deepheart, mehrgardt2021deep}. Despite significant progress, most existing studies primarily focus on ID testing, where the train and test datasets stem from the same distribution. This approach does not consider OOD evaluation scenarios and thus does not reflect the reality of real-world applications, where test data can come from various distributions. Real-world test sets might differ from the training dataset in various aspects, such as BP distributions, sensor hardware used for capturing, signal quality\added{,} and subject physiologies. 

This study aims to address these gaps by leveraging the PulseDB dataset \cite{wang2023pulsedb} as a training dataset, and then evaluating both ID and OOD generalization of DL models for the estimation of BP \added{from the PPG signal alone}. The goal is to identify model architectures and training datasets that \replaced{allow to train models with robust OOD performance}{give rise to models that show a robust OOD performance}. The PulseDB dataset was used for training due to its large size. Then, four external datasets served as test datasets to assess OOD generalization. Furthermore, we investigated a simple domain adaptation approach to improve OOD generalization. \added{ This paper does not propose a new domain adaptation algorithm. Instead, it explores how adapting to the target dataset's label distribution affects which datasets or training scenarios are most effective for developing models with robust out-of-distribution performance.}  This work offers a comprehensive benchmarking analysis on diverse datasets, providing insights into the robustness of these models in real-world scenarios. Finally, we closed with practical recommendations on model architectures, training datasets, and scenarios to achieve good OOD generalization.

\section{Related Work}
\label{sec:II}

\heading{Challenges for BP estimation and benchmarking models} 
\label{sec:II-B}
Several research studies have indicated that the effectiveness of learning models in predicting BP is highly dependent on the quality and quantity of the training data \cite{qin2023machine, treebupachatsakul2022cuff, zabihi2021bp}. \deleted{For instance,  researchers who worked with the MIMIC-III dataset underscore the significance of data preparation to achieve better model outcomes \citex{chu2023non}.} The aforementioned challenges result in a growing demand for reliable benchmark studies to thoroughly evaluate different ML/DL methods for BP estimation using PPG, utilizing comprehensive datasets to measure their efficiency. Consequently, benchmark investigations have been performed in this area to address the aforementioned demands. We refer to \cite{gonzalez2023benchmark} as a notable and recent benchmark study that used four different datasets. 
However, the mentioned study only considers models trained from scratch on the respective datasets and evaluates them on ID test sets, which are known to provide overly optimistic measures for the generalization performance on unseen data. The PulseDB dataset, a large-scale, high-quality dataset containing PPG signals and reference BP measurements\deleted{,} and \replaced{is therefore}{as such is} a unique resource for training deep-learning based BP prediction models, after which they can be externally validated on external datasets.

\heading{Challenges of OOD generalization} 
Most of the ML/DL techniques \replaced{typically}{usually} rely on the subtle statistical patterns that may exist within the training data, hence functioning under ideal conditions where both the training and testing data belong to the same distribution (ID). However, this perfect situation rarely occurs in real-world scenarios \cite{zhang2021deep, engstrom2019exploring}. \replaced{Prior studies have}{Previous work has} shown that most DL models perform poorly on tasks induced by data from distributions other than their training data (OOD) generalization \cite{ballas2022domain}. The concept of OOD generalization and its application to DL models has evolved with contributions from various researchers, e.g., \cite{gwon2023out, yi2021improved, ye2021towards}. 
The challenges of OOD generalization in the context of PPG-based BP estimation have been investigated in a recent publication \cite{weber2023intensive}. They focused on feature-based approaches, whereas the present work covers deep learning models operating on raw time series. This work \replaced{aims to provide a more comprehensive understanding}{establishes a more comprehensive picture} by considering a large number of external datasets and investigating the potential impact of domain adaptation.

\heading{Improving OOD generalization} 
It is worth mentioning that OOD generalization is a challenging task that may result in poor performance\deleted{s} since unseen data very often do not resemble the training set \cite{hendrycks2017baseline, liang2018enhancing}. Several previous works have intensively addressed the challenge of mitigating the influence of OOD signals \replaced{in ECG and EEG data analysis.}{when analyzing ECG and EEG data.} \cite{ballas2022domain,soltanieh2023distribution} have shown the effectiveness of using domain generalization and self-supervised learning approaches to improve the classification accuracy of OOD ECG signals. Also, recent work by \cite{yang2022multimodal} has demonstrated promising results in addressing domain shifts and OOD signals between diverse EEG datasets, \replaced{highlighting the potential of domain adaptation to enhance EEG signal recognition robustness.}{highlighting the potential of domain adaptation techniques to enhance the robustness of EEG signal recognition}. According to previous studies, one of the key approaches to tackle the challenge of OOD generalization is to reduce the influence of distribution shifts between training and test sets by revising the distribution of training data to mimic the distribution of test data, aiming to minimize the predictive error on the test set \cite{ben2010theory,bickel2009discriminative}.
In this work, we used a simple sample-based empirical risk minimization approach based on sample weights inferred from the label distribution (i.e., BP reference labels) in the source and target domains to assess the potential benefits of incorporating domain adaptation approaches. 

\heading{Technical contributions} 
In this work, we put forward the following technical contributions:

\begin{enumerate}

\item We implemented state-of-the-art DL-based time series classification algorithms for PPG-based BP estimation on the large-scale, high-quality PulseDB dataset and evaluated their performance in a first comprehensive comparative study. 

\item We investigated both ID and OOD generalization of models trained on various PulseDB subsets. These models were evaluated on different PulseDB subsets and four external datasets. To contextualize OOD performance, we compared it with the differences in label distributions between the training and test datasets.

\item We assessed the benefit of domain adaptation by using an importance-weighted empirical risk minimization approach using importance weights \replaced{inferred from the respective label distributions, and we put forward}{inferred from the respective label distributions and put forward} recommendations for training dataset choices that promise good generalization properties.

\end{enumerate}

\section{Materials \& Methods}
\label{sec:III}

\subsection{Training and Evaluation Datasets}
\label{sec:III_A}

\begin{table*}[ht]
\centering
\caption{Summary of the datasets utilized in this study: Two subsets PulseDB \cite{wang2023pulsedb} for training and four external datasets \cite{gonzalez2023benchmark} for OOD evaluation.}
\scalebox{0.80}{
\renewcommand{\arraystretch}{1.2}
\begin{tabular}{lccccccc}

& \multicolumn{2}{c}{\textbf{PulseDB}} & \multicolumn{4}{c}{\textbf{External Datasets}} \\ 
\cmidrule(lr){2-3} \cmidrule(lr){4-7}
\textbf{Metric} & \textbf{MIMIC} & \textbf{VitalDB} & \textbf{Sensors} & \textbf{UCI} & \textbf{BCG} & \textbf{PPGBP} \\
\midrule
\textbf{Subjects} & 1,474 & 1,553 & 1,195 & \replaced{10,793}{unknown} & 40 & 218 \\
\textbf{Total Duration (h)} & $\sim$2357 & $\sim$1793 & $\sim$15 & $\sim$570 & $\sim$4 & $<$1 \\
\textbf{Segments (number , length)} & 848,796 , 10s   & 645,678 , 10s & 11,102 , 5s & 410,596 , 5s & 3,063 , 5s & 619 , 2.1s \\
\textbf{Age (mean ± SD)} & 61.16 ± 15.29 & 58.71 ± 15.14 & 57.1 ± 14.2 & unknown & 34.2 ± 14.5 & 56.9 ± 15.8 \\
\textbf{Gender} & 45.94\% F, 54.05\% M & 41.84\% F, 58.15\% M & 40.2\% F, 59.8\% M & unknown & 55.5\% F, 44.5\% M & 53.1\% F, 46.9\% M \\
\textbf{SBP (mmHg, mean ± SD)} & 123.32 ± 23.00 & 115.62 ± 18.92 & 134.36 ± 21.78 & 131.57 ± 11.16 & 120.99 ± 15.29 & 128.02 ± 20.50 \\
\textbf{DBP (mmHg, mean ± SD)} & 61.58 ± 13.48 & 63.03 ± 12.05 & 65.37 ± 10.51 & 66.79 ± 10.48 & 67.23 ± 9.30 & 71.91 ± 11.20 \\
\bottomrule
\end{tabular}
}
\label{tab:tab1}
\end{table*}

\begin{table*}[h]
\centering
\caption{Summary of generated PulseDB subsets (Samples / Subjects)}

\begin{tabular}{llcccc}
\textbf{Source} & \textbf{Subset} & \textbf{Train} & \textbf{Validation} & \textbf{Calibration} & \textbf{Test} \\
\midrule
\multirow{3}{*}{\textbf{Combined}} 
& \textbf{Calib} &  811,955 / 2,494 & 78,899 / 2,494 & 11,306 / 2,494 & 100,240 / 2,494 \\
& \textbf{CalibFree} & 801,720 / 2,217 & 66,960 / 186 & 33,480 / 93 & 111,600 / 279 \\
& \textbf{AAMI} & 902,160 / 2,494 & 230,145 / 149 & 148,989 / 93 & 1340 / 242 \\
\midrule
\multirow{3}{*}{\textbf{Vital}} 
& \textbf{Calib} & 418,986 / 1,293 & 40,673 / 1,293 & 5821 / 1,293 & 51720 / 1,293 \\
& \textbf{CalibFree} & 416,880 / 1,158 & 32,400 / 90 & 1000 / 45 & 57,600 / 144 \\
& \textbf{AAMI} & 465,480 / 1,293 & 43,820 / 71 & 26,392 / 45 & 666 / 116 \\
\midrule
\multirow{3}{*}{\textbf{MIMIC}} 
& \textbf{Calib} & 392,969 / 1,213 & 38,226 / 1,213 & 5,485 / 1,213 & 48,520 / 1,213 \\
& \textbf{CalibFree} & 384,840 / 1,069 & 34,560 / 96 & 17,280 / 48 & 54,000 / 135 \\
& \textbf{AAMI} & 436,680 / 1,213 & 186,325 / 78 & 122,597 / 48 & 674 / 126 \\
\bottomrule
\end{tabular}
\label{tab:tab2}
\end{table*}

\heading{PulseDB dataset} 
PulseDB is sourced from selected pre-processed signals from the MIMIC-III \cite{moody2020mimic} and VitalDB \cite{lee2022vitaldb} databases. It is one of the most extensive datasets currently available, containing 5,245,454 10-second segments of ECG, PPG, and arterial BP (ABP) waveforms across 5,361 subjects. The dataset includes demographic details such as age, gender, weight, height, and body mass index (BMI). Both VitalDB and MIMIC-III represent samples collected from finger-tip PPG sensors from patients undergoing surgery and in Critical Care Units, respectively.
It is worth noting that the PulseDB dataset is categorized into the following subsets, making it ideal for benchmarking cuff-less BP estimation models:
\begin{itemize}
    \item \textit{Calib:} Created for a calibration-based approach, where each subject contributes data to both the training and testing sets. This enables the model to adapt to patient-specific signal features in order to improve the prediction performance. The focus of this scenario is to train models that show good generalization to unseen samples of patients encountered during training.
    \item \textit{CalibFree:} Created for a calibration-free approach, in which training and test sets do not share any subjects. The focus of this scenario is to develop models that generalize to entirely unseen patients.
    \item \textit{AAMI:} Created for a second calibration-free scenario, which complies with the high standards developed by the Association for the Advancement of Medical Instrumentation (AAMI) \cite{stergiou2018universal}. The main difference between this and the CalibFree scenario is a stronger emphasis on the tails of the BP distribution. The focus of this scenario is to assess the generalization to unseen patient\added{s} with the stricter protocol of the AAMI for medical device testing. 
\end{itemize}

We generated nine subsets of the PulseDB dataset, inspired by the instructions in the original PulseDB publication \cite{wang2023pulsedb} and the corresponding code repository. Tables \ref{tab:tab1} and \ref{tab:tab2} summarize the utilized PulseDB dataset in this paper (including data from MIMIC and VitalDB), and the generated subsets, respectively. We have three major subsets, Calib, CalibFree, and AAMI, which are derived from VitalDB or MIMIC, or a combination of VitalDB and MIMIC (combined) sources, resulting in nine different subsets.  We kept the original test sets intact to ensure comparability with results in the literature; however, we split off additional validation and calibration sets from the respective training sets, mimicking the way in which the \replaced{corresponding}{respective} test sets were constructed.

\heading{External datasets} We used the datasets presented in a recent benchmark study \cite{gonzalez2023benchmark} as external datasets, as they are qualitatively very different \replaced{from}{to} PulseDB in terms of sample size, signal quality\added{,} and patient \replaced{population}{collective}, and are therefore well-suited to investigate the OOD generalization of models trained on PulseDB or subsets thereof. \cite{gonzalez2023benchmark} provides \deleted{the} pre-processed versions of each external dataset along with \replaced{the corresponding}{their} pre-processing methods, which \replaced{we use}{served} as the external \replaced{in this}{datasets for the purpose of our} study (Table \ref{tab:tab1}). These datasets are comprehensively described in \cite{gonzalez2023benchmark}. Therefore, we refer to the original publication for details, and now briefly outline the key features of each dataset:

\begin{itemize}
\item \textit{Sensors}: The Sensors dataset is derived from MIMIC-III with simultaneous PPG and ABP waveforms from 1,195 ICU patients. After pre-processing, it includes two 15-second segments per record with a 5-minute interval between them. 

\item\textit{UCI}: The UCI dataset is derived from the MIMIC-II dataset and is the largest \replaced{of}{dataset in} our external dataset\added{s}.

\item\textit{BCG}: The BCG dataset contains recordings from 40 subjects (primarily healthy). Although the BCG is a smaller dataset after pre-processing with limited variability, the number of segments per subject is remarkably high. 

\item\textit{PPGBP}: The PPGBP dataset contains data from 219 subjects, each with cardiovascular conditions. Following pre-processing, PPGBP becomes the shortest dataset, containing 218 subjects and 613 segments of 2.1s length, making PPGBP the smallest dataset in terms of total duration. 
\end{itemize}

\subsection{Prediction Models} 
\label{sec:III_B}

\heading{Overview of considered model architectures}

The number of DL approaches for time series classification is immense, and we refer to \cite{ mohammadi2024deep, liang2024foundation} for an extensive review of these methods. Since our work pertains to BP estimation, \replaced{in which}{wherein} signal processing plays a major role, we leveraged these foundations by evaluating several convolutional neural network architectures (CNN). Also, in this paper, we extended our exploration to include structured state space sequence (S4) models \cite{guefficiently}, which are known for their ability to effectively capture long-range dependencies, and \replaced{have shown}{showed} promising results for other physiological time series \cite{mehari2023towards,wang2023s4sleep,saab2024towards, wang2024assessing}.

CNNs have been part of time series analysis for a long time by offering flexibility and scalability in model design. In this work, we evaluated the performance of three main CNN architectures: 

\heading{Simple feed-forward CNN architectures} 
The most straightforward convolutional neural network is a neural network without cycles since it includes data flow only in one direction through its layers. A prototypical example of such architecture is the LeNet1D \cite{lecun1998gradient}. Our work builds on the one-dimensional adaptation put forward in \cite{wagner2024explaining}. 

\heading{ResNet-based Architectures} 
The ResNet model has made a crucial stride in DL by proposing skip connections, which enable\deleted{d} \replaced{easier}{easy} gradient flow via backpropagation. Here, we drew on one-dimensional ResNet variants, such as XResNet1d50 and XResNet1d101, proposed in \cite{strodthoff2020deep}. 

\heading{Inception-based Architectures} 
Inception models, originally proposed in computer vision \cite{szegedy2015going}, include several convolutional filters with different kernel sizes to capture a broader spectrum of feature patterns. Furthermore, such a hierarchical feature extraction approach has benefited physiological signal analysis, where intricate\added{,} generalized patterns \replaced{need}{are} to be captured for proper interpretation. Specifically, several previous studies have applied Inception1D \cite{ismail2020inceptiontime} in time series classification tasks and reported its excellent performance owing to its \replaced{strong}{good} \replaced{representational}{representation} capability with comprehensive and diverse features \cite{strodthoff2020deep, ismail2020inceptiontime}.

\heading{Structured State Space Sequence (S4) Models}
As an alternative model category, we considered a structured state space sequence model, which has been successfully applied to physiological time series \cite{wang2023s4sleep, mehari2023towards} and is known for its ability to capture long-range dependencies in input sequences \cite{gu2021efficiently}. Here, we used the S4 model as a prediction model as in \cite{mehari2023towards}.

\subsection{Training and evaluation procedures}
\label{sec:III-C}
\heading{Training procedure}

For each of the experiments, an effective batch size of 512 was used through gradient accumulation. The learning rates were \deleted{either} found using a learning rate finder \cite{smith2017cyclical} \added{for S4 model} and set to 0.001 \added{for other models}. Models were trained for 50 epochs. \replaced{T}{Also, t}he training \deleted{routine} was \replaced{performed}{implemented} \replaced{using}{with} the AdamW optimizer \cite{loshchilov2017decoupled} and mean squared error as loss function. In all cases, consistent with previous studies, e.g., \cite{xiao2024advancing}, we employed two output nodes to jointly predict SBP and \replaced{DBP}{DPB}, leveraging possible shared physiological features to enhance model performance \replaced{for both}{among} SBP and DBP to improve model performance. \deleted{We used the validation set score for hyperparameter tuning.} As a simple measure to reduce overfitting, we performed model selection based on the validation set score, i.e., during training, we kept track of the validation set score and selected the model with the best validation set score for evaluation on ID or OOD test sets.

All of the considered datasets used a sampling frequency of 125~Hz. \replaced{Models were trained using the full input resolution of each dataset: 1250 time steps (10 seconds) for PulseDB, 625 time steps (5 seconds) for the BCG, UCI, and Sensors datasets, and 262 time steps (2.1 seconds) for PPGBP. The fact that all considered model architectures involve a global average-pooling allows for flexible input lengths at inference time, while still leveraging the model trained on PulseDB.}{Models were trained using the samples' full input resolution, i.e., 1250 time steps in the case of PulseDB. The fact that all considered model architectures involve a global average-pooling, allows to evaluate models also for other input sizes, such as the native input sizes of the external datasets (625 time steps for BCG, UCI, and Sensors datasets, and 262 time steps for PPGBP) while still leveraging the model trained on PulseDB.}

\heading{\added{Primary} performance metrics}
As the primary performance metric referenced by BP standards  \cite{iso2018sphygmomanometers, IEEE1708a2019}, we reported the mean absolute error (MAE) defined by 

\begin{equation}
\text{MAE} = \frac{1}{n} \sum_{i=1}^{n} |\text{Predicted}_i - \text{Reference}_i|,
\end{equation}

where \(n\) is the number of predictions, \(i\) is the index, \(\text{Predicted}_i\) and \(\text{Reference}_i\) indicate the predicted and reference BP, respectively. 

\added{\heading{Statistical significance} We used empirical bootstrapping on the test set to shed light on the statistical fluctuations due to the finiteness and composition of the respective test sets. For each investigated scenario, we selected the best-forming model, i.e., the model with the lowest MAE as reference. We applied bootstrapping to the performance difference between the reference model and every other model under consideration using 1,000 bootstrap iterations. The other model is considered statistically significantly worse if the estimated 95$\%$ confidence intervals contain 0. Otherwise, we considered the other model as not statistically significantly worse than the reference model and mark the model accordingly in the results tables.}

\added{\heading{Secondary performance metric} To provide deeper insights into the nature of the disagreement between the predictions, we performed Bland–Altman analyses \cite{bland1986statistical}. To this end, we reported the bias (mean of the differences between prediction and reference) and its limits of agreement (LoA; 1.96 times the standard deviation of the differences). As compared to MAE, the bias provides insights into systematic deviations between prediction and reference. The LoA reveals the spread of the individual differences and not just averages.}

\subsection{A simple baseline for domain adaptation}
\label{sec:III_D}

\heading{Approaches to domain adaptation} The challenge of dealing with an inevitable mismatch between training and test set distributions is a long-standing one and \added{has} attracted a lot of interest in the machine learning community, see \cite{kouw2019review} for a recent review. Conventionally, \replaced{researchers}{one} distinguish\deleted{es} \added{between} sample-based, feature-based, and inference-based approaches. In this work, we explored the potential benefit of domain adaptation methods in a simple sample-based approach, where we deviated from the paradigm of target-free domain adaptation through the use of the target domain label distribution. Most importantly, we only made use of the target domain label distribution but not of individual labels, which represents a piece of information that we envisioned to be typically available in practical use cases. The main motivation is to assess the potential benefit of domain adaptation on the model performance.

\begin{figure}[t]  
  \centering
  \includegraphics[width=\columnwidth]{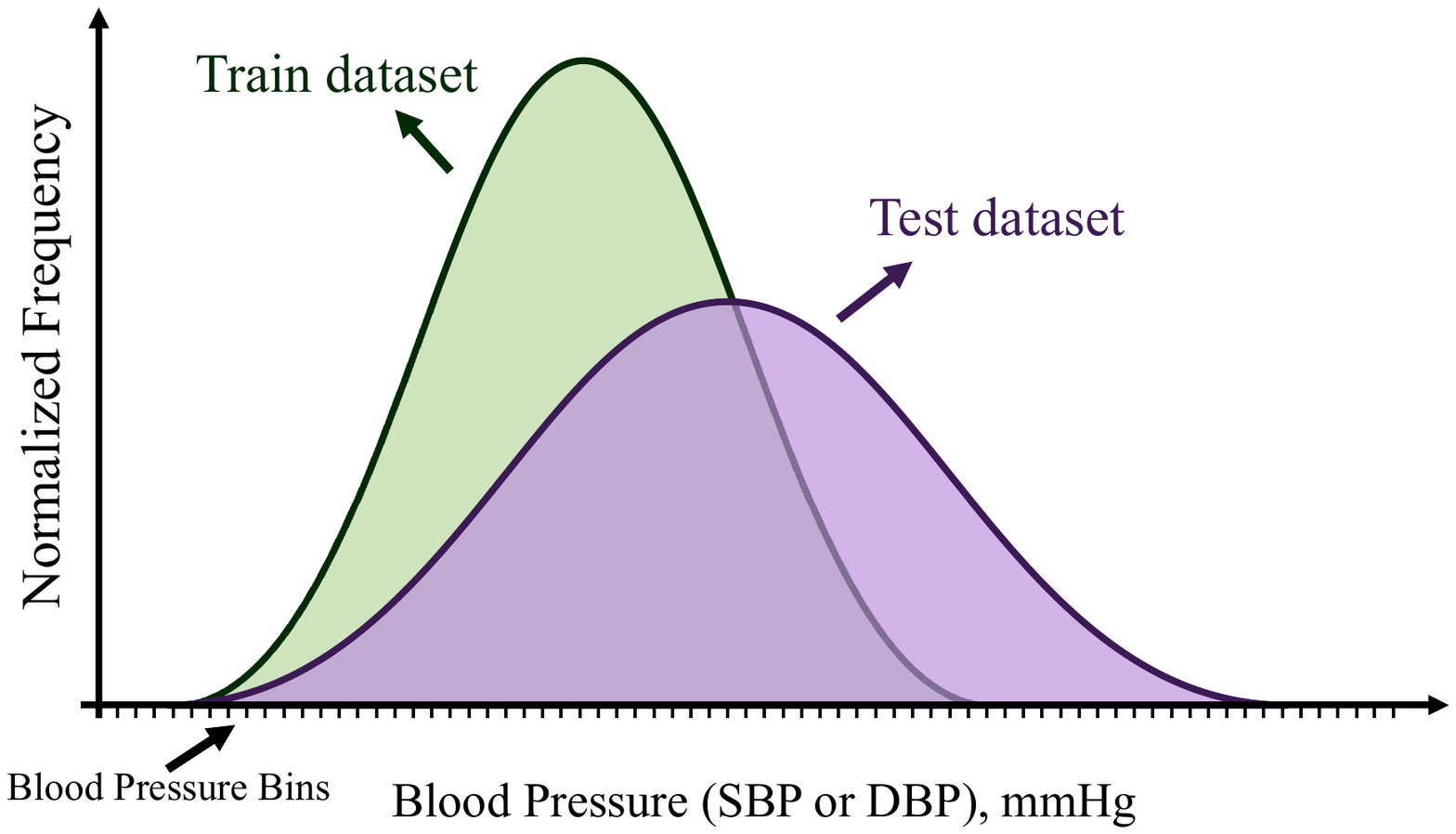}
  \caption{Schematic comparison of normalized label distributions (e.g., SBP or DBP) across training and test datasets.}
  \label{fig:fig5}
\end{figure}

\heading{Reweighting based on label distributions} More specifically, we proposed to use an empirical risk minimization approach using sample weights derived from the difference \replaced{between}{of} the label distributions of the respective source domain and target domain test datasets. To this end, we summarized both label distributions in terms of (normalized) histograms, see Figure~\ref{fig:fig5}. For a given training sample that belongs to the bin $i$, we identified the empirical output probability $h_{\text{train},i}$($h_{\text{test},i}$) assessed from the corresponding test set histograms. \replaced{Based on this}{From that}, we defined sample weights via
\begin{equation}
w_i = 
\begin{cases} 
\max(\added{\tau}, \frac{h_{\text{test},i}}{h_{\text{train},i}}) & \text{if } h_{\text{train},i} > 0 \\
\added{\tau} & \text{if } h_{\text{train},i} = 0\,,
\end{cases}
\label{eq:weight_calculation}
\end{equation}
where the hyperparameter $\tau$ is used to prevent excluding training samples for which the relative weight $h_{\text{test},i}/h_{\text{train},i}$ is small entirely from the training process. In our experiments, we fixed $\tau=1$, but found that the results (validation set scores) were not very sensitive to the choice of $\tau$.

Finally, we utilized the sample weights $w_i$ for the loss calculation. More specifically, we used importance weights derived from the SBP distribution for the loss calculated based on the SBP output and importance weights derived from the DBP distribution for the loss calculated based on the DBP output and sum both contributions to obtain the final importance weighted loss.

\section{Results}
\label{sec:IV}

\heading{Organization of the experiments} In Section \ref{sec:IV-A}, we evaluated classifiers for all nine generated PulseDB subsets, which \replaced{forms}{is} the core of our analysis. Section \ref{sec:IV-B} supplements the above analysis by \replaced{evaluating}{performing} ID and OOD generalizations within PulseDB datasets. Also, the OOD generalization on external datasets is investigated in Section \ref{sec:IV-C}. Finally, in Section \ref{sec:IV-D}, we conducted similar analyses as in Sections \ref{sec:IV-B} and \ref{sec:IV-C}, however, using the importance weighting for domain adaptation. We computed and evaluated the results with a focus on reducing the distribution shift between the training and test sets by incorporating a \replaced{specific}{special} weight value for each training sample.

\subsection{Model comparison on PulseDB}
\label{sec:IV-A}

\begin{figure*}[htbp] 
\centering

\includegraphics[width=1\textwidth]{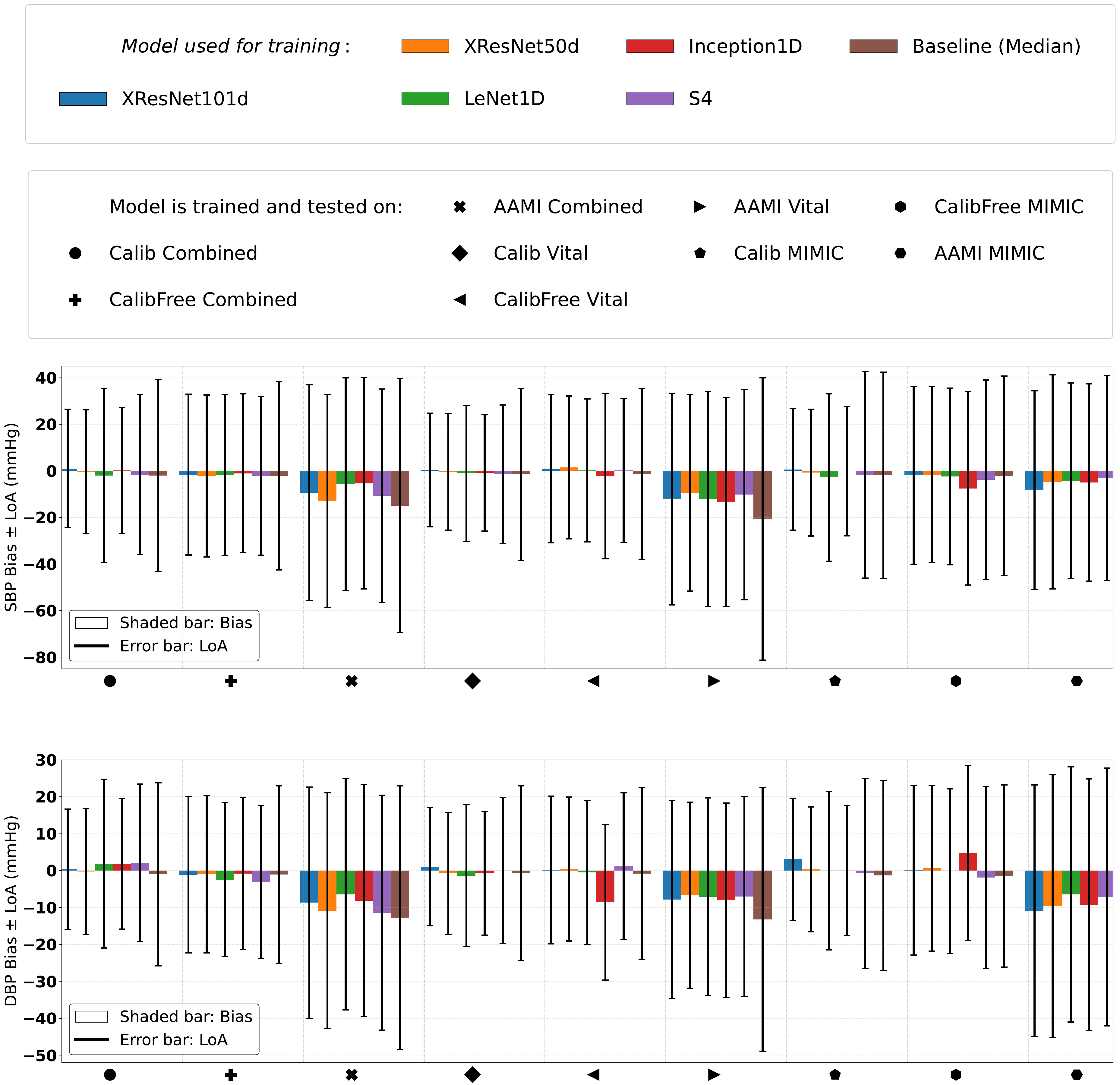} 
\caption{\added{Prediction bias (mean error) and LoA of different models trained and tested on different PulseDB subsets for SBP (top) and DBP (bottom). Corresponding MAE values are indicated in Table \ref{tab:tab3}.}}
\label{fig:figure1}
\end{figure*}

\heading{Overview} We carried out all experiments for the nine subsets in Table \ref{tab:tab1} using all models presented in Section \ref{sec:III_B}. \added{For each subset, models were trained, validated, and evaluated using the corresponding training, validation, and independent test sets as defined in Table \ref{tab:tab2}. In the CalibFree and AAMI settings, there is no subject overlap between training and test sets}. \replaced{The results of these experiments are compiled}{All experiment results (SBP/DBP) are shown} in Table \ref{tab:tab3} and Figure \ref{fig:figure1}. 

\added{In Table \ref{tab:tab3}, models with the lowest MAE and no statistically better alternative (based on bootstrap analysis  as described in section \ref{sec:III-C}) are shown in bold for each experiment.} The best-ranked models, either based on ResNet or Inception architectures, achieved MAEs between \added{about} 9 mmHg (6 mmHg) in the Calib Vital subset to approximately 12 mmHg (8 mmHg) in the CalibFree Vital and \replaced{around}{up to} 18 mmHg (13 mmHg) in the AAMI tasks for the Combined subset for systolic (diastolic) blood pressure. These results already provide first insights into the relative complexity of the different prediction tasks across all considered scenarios. \replaced{As expected}{Not surprisingly}, the scores achieved for the Calib tasks are comparably lower (compared to CalibFree and AAMI tasks) as the model can exploit subject-specific information about test set samples through samples from the corresponding subjects seen during training.

\begin{table*}[ht]
\centering
\caption{Performance of PPG DL models on PulseDB dataset in terms of MAE (SBP / DBP) measured in units of mmHg. \deleted{For each of the experiments, the three best-performing models are marked in bold-face and the overall best-performing model is underlined.} \added{For every setting, we underlined the best-performing model and boldface models that do not perform statistically significantly worse than this model.}}

\scalebox{0.8}{
\renewcommand{\arraystretch}{1.2}
\begin{tabular}{lccccccccc}
\multicolumn{1}{c}{} & \multicolumn{9}{c}{\textbf{Model is trained and tested on}} \\
\cmidrule(lr){2-10}
 & \multicolumn{3}{c}{\textbf{Combined}} & \multicolumn{3}{c}{\textbf{Vital}} & \multicolumn{3}{c}{\textbf{MIMIC}}  \\ 
\cmidrule(lr){2-4} \cmidrule(lr){5-7} \cmidrule(lr){8-10}
\textbf{\textit{Model used for training}} & 
\textbf{Calib $\downarrow$} & \textbf{CalibFree $\downarrow$} & \textbf{AAMI $\downarrow$} 
& \textbf{Calib $\downarrow$} & \textbf{CalibFree $\downarrow$} & \textbf{AAMI $\downarrow$} 
& \textbf{Calib $\downarrow$} & \textbf{CalibFree $\downarrow$} & \textbf{AAMI $\downarrow$}  \\ 
\midrule

\textbf{\textit{Baseline (Median)}} & 16.66 / 9.85 & 16.48 / 9.75 & 25.48 / 17.29 & 14.92 / 9.52 & 14.88 / 9.44 & 29.85 / 17.84 & 18.16 / 10.07 & 17.69 / 9.95 & 21.23 / 16.82  \\ 
\textbf{\textit{Lenet1D}} & 14.66 / 9.20 & 13.88 / 8.52 & \textbf{18.57 / \underline{13.36}} & 11.61 / 7.70 & \textbf{\underline{12.37}} / 7.89 & 19.59 / 11.87 & 14.37 / 8.22 & 15.41 / \underline{\textbf{8.92}} & \textbf{17.56 / 14.51}  \\ 

\textbf{\textit{XResNet1d50}} & 9.96 / 6.35 & 14.12 / 8.57 & 20.49 / 15.11 & 9.49 / 6.33 & 12.40 / \textbf{\underline{7.85}} & \underline{\textbf{17.71 / 11.43}} & 10.14 / \textbf{\underline{6.18}} & \underline{\textbf{15.36}} / 9.09 & 19.02 / 15.94  \\ 
\textbf{\textit{XResNet1d101}} & \underline{\textbf{9.43 / 5.98}} & 13.97 / 8.51 & 19.38 / 14.04 & \underline{\textbf{9.09 / 6.09}} & 12.70 / 8.05 & 19.31 / 12.33 & \textbf{\underline{9.52}} / 6.64 & 15.47 / 9.27 & 18.35 / 15.65 \\ 

\textbf{\textit{Inception1D}} & \textbf{10.37 / 6.98} & \underline{\textbf{13.71 / 8.27}} & \underline{\textbf{18.21}} / 13.83 & 9.65 / 6.53 & 14.54 / 10.96& 19.79 / 12.30 & 10.52 / 6.52 & 17.46 /10.29 & \underline{\textbf{17.33}} / 15.00 \\ 
\textbf{\textit{S4}} & 13.65 / 8.66 & 13.76 / 8.62 & 19.57 / 15.43 & 11.92 / 7.91 & \textbf{12.39} / 8.03 & 18.40 / 12.19 & 13.16 / 10.08 & 17.79 / 9.96 & \textbf{17.83} / 14.88 \\
\bottomrule
\end{tabular}
}
\label{tab:tab3}
\end{table*}


\heading{Best-performing models} \replaced{Assessing the overall model performance by the number of boldface entries in Table~\ref{tab:tab3}, the three deep neural networks XResNet1d50, XResNet1d101 and Inception1D, but also the shallow LeNet show a comparable level of performance.}{Aggregating model performance by counting best-performing models across all different setups, we concluded that ResNet and Inception-based frameworks generally represent the best-performing models.} \replaced{In particular, the}{The shallow} Lenet1D models \replaced{demonstrate relatively strong performance within}{are competitive in} the CalibFree category and partly also in the AAMI category, but fail to achieve \replaced{similarly strong}{competitive} results in the Calib scenario, which \replaced{appears to benefit}{profits} to a certain degree from model\added{'s} capacity to memorize specific patients. For instance, according to Table \ref{tab:tab3}, Lenet1D is positioned in the same range as state-of-the-art architectures such as ResNet and Inception in distinct subsets (e.g., CalibFree Combined, CalibFree Vital, AAMI Combined, and AAMI MIMIC). \deleted{e.g., on the subset of CalibFree Combined, Lenet1D has a MAE of 13.88 mmHg (8.52 mmHg), which is very close to XResNet1d101 performance with 13.97 mmHg (8.51 mmHg), and outperforms XResNet1d50 with 14.12 mmHg (8.56 mmHg).} This means that Lenet1D, although \replaced{a}{the} simpler architecture, can provide robust results for the estimation of BP \textit{in certain scenarios}, as similarly reported in \cite{wagner2024explaining} for ECG analysis. 
Furthermore, it is worth noting that the S4 model does not show the outstanding performance it demonstrated in the ECG/EEG domain \cite{mehari2023towards,wang2023s4sleep,saab2024towards}.
\deleted{As a final general observation, throughout most subsets, XResNet1d101 consistently ranks among the top-performing models. Therefore, we selected XResNet1d101 as the DL model for the further sections of this paper.} \added{To simplify the following analysis, we selected a single model architecture, the XResNet101d, that shows competitive performance across all prediction scenarios for the following investigations.} A recent performance comparison among feature-based, image-based, and raw-time-series-based models confirmed the superiority of raw-time-series-based models, as demonstrated in \cite{qumphyd1}.

\added{\heading{Bland-Altman analysis} Turning to Figure ~\ref{fig:figure1}, the Bland-Altman analysis results reveal that all AAMI tasks have large negative biases compared to the other test subsets, typically exceeding a bias of 10 mmHg, and in some models, even beyond 15 mmHg. Conversely, most of the other test subsets exhibit biases closer to zero, generally within the range of ±5 mmHg. This discrepancy is probably due to the dissimilarity between the blood pressure distributions of the respective training datasets and the AAMI test sets. This hypothesis is supported by the fact that the baseline models show a bias of comparable size. It can also be read off from the blood pressure distributions in Figures \ref{fig:Grouped_BP_SBPDBP_Train_RefCalibFreeVital} and \ref{fig:Grouped_BP_SBPDBP_TestOnly_RefCalibFreeVital}, which are biased towards higher blood pressure values. The bias, in combination with the corresponding LoA, indicates that the purely PPG signal-based blood pressure estimation does not agree closely with the reference BP measurement. Even though the Bland-Altman analysis provides interesting insights into systematic deviations of the predictions, we focused on MAE as the primary metric in the following sections, in line with previous studies \cite{wang2023pulsedb,gonzalez2023benchmark}, and industry standards \cite{IEEE1708a2019}. Corresponding Bland-Altman analysis for the results presented in the following sections can be found in the supplementary materials.}

\subsection{Performance evaluation within PulseDB}
\label{sec:IV-B}

In this section, we presented the performance of our selected model, XResNet1d101, trained and tested on various subsets of PulseDB. The results are shown in Table \ref{tab:tab4}, where the train\added{ing} and test sets are represented in the vertical and horizontal columns, respectively. We organized the results by data source (Combined/Vital/MIMIC) and scenario (Calib/CalibFree/AAMI) in order to provide a comprehensive analysis of the results.

\begin{table*}[htbp]
\centering
\caption{Performance of ID and OOD generalization on all subsets of PulseDB dataset for an XResNet1d101 model in terms of MAE (SBP / DBP) given in units of mmHg. Vertical and horizontal subsets represent the train\added{ing} and test sets, respectively. \added{For every setting, we underlined the best-performing model and boldface models that do not perform statistically significantly worse than this model.}} 

\scalebox{0.80}{
\renewcommand{\arraystretch}{1.2}
\begin{tabular}{llccccccccc}
 & & \multicolumn{9}{c}{\textbf{Model is tested on}} \\
\cmidrule(lr){3-11}
 & & \multicolumn{3}{c}{\textbf{Combined}} & \multicolumn{3}{c}{\textbf{Vital}} & \multicolumn{3}{c}{\textbf{MIMIC}} \\
\cmidrule(lr){3-5} \cmidrule(lr){6-8} \cmidrule(lr){9-11}
\multicolumn{2}{l}{\hspace{2.5em}\textbf{\textit{Model is trained on}}} & \textbf{Calib $\downarrow$} & \textbf{CalibFree $\downarrow$} & \textbf{AAMI $\downarrow$} & \textbf{Calib $\downarrow$} & \textbf{CalibFree $\downarrow$} & \textbf{AAMI $\downarrow$} & \textbf{Calib $\downarrow$} & \textbf{CalibFree $\downarrow$} & \textbf{AAMI $\downarrow$} \\
\midrule
\multirow{3}{*}{\textbf{\textit{Combined}}}
& \textbf{\textit{Calib}} & \textbf{\underline{9.43} / \underline{5.98}} & 15.53 / 9.28 & \textbf{\underline{19.20}} / 14.22 & 9.28 / \textbf{\underline{6.08}} & 13.97 / 8.65 & 18.57 / 12.46 & \textbf{\underline{9.59} / \underline{5.86}} & 17.20 / 9.95 & 19.82 / 15.97 \\
& \textbf{\textit{CalibFree}} & 13.87 / 8.53 & \textbf{13.97 / \underline{8.51}} & 20.10 / 14.73 & 12.34 / 8.23 & 12.58 / 8.10 & 21.75 / 14.42 & 15.51 / 8.85 & 15.45 / 8.95 & \textbf{18.47} / 15.04 \\
& \textbf{\textit{AAMI}} & 13.61 / 8.50 & \textbf{\underline{13.96} / 8.52} & \textbf{19.38} / 14.04 & 12.19 / 8.12 & 12.56 / 8.02 & 20.47 / 13.60 & 15.12 / 8.90 & 15.45 / 9.04 & \textbf{\underline{18.30} / \underline{14.47}} \\
\midrule
\multirow{3}{*}{\textbf{\textit{Vital}}} 
& \textbf{\textit{Calib}} & 14.69 / 9.74 & 16.71 / 10.80 & 21.32 / 13.95 & \textbf{\underline{9.09} / 6.09} & 13.92 / 8.74 & 19.17 / 11.99 & 20.67 / 13.63 & 19.68 / 12.99 & 23.45 / 15.88 \\
& \textbf{\textit{CalibFree}} & 15.37 / 8.82 & 16.37 / 9.20 & 20.21 / \textbf{\underline{13.30}} & 10.90 / 7.25 & 12.70 / 8.05 & \textbf{\underline{17.46} / \underline{11.55}} & 20.13 / 10.50 & 20.29 / 10.43 & 22.93 / 15.02 \\
& \textbf{\textit{AAMI}} & 15.12 / 9.13 & 14.82 / 8.89 & 20.22 / \textbf{13.44} & 11.84 / 7.91 & \textbf{\underline{12.18} / \underline{7.87}} & 19.31 / 12.33 & 18.61 / 10.43 & 17.64 / 9.98 & 21.11 / \textbf{14.53} \\
\midrule
\multirow{3}{*}{\textbf{\textit{MIMIC}}} 
& \textbf{\textit{Calib}} & 12.91 / 8.63 & 16.60 / 10.34 & 22.94 / 16.27 & 16.08 / 10.49 & 16.24 / 10.27 & 26.05 / 17.04 & \textbf{9.52} / 6.64 & 16.99 / 10.42 & 19.87 / 15.50 \\
& \textbf{\textit{CalibFree}} & 14.69 / 9.03 & 15.16 / 9.28 & 21.36 / 15.86 & 14.24 / 9.12 & 14.87 / 9.30 & 24.05 / 16.68 & 15.16 / 9.28 & 15.47 / 9.27 & \textbf{18.71} / 15.05 \\
& \textbf{\textit{AAMI}} & 14.23 / 9.29 & 14.59 / 9.47 & 21.54 / 16.98 & 13.70 / 9.96 & 14.20 / 10.03 & 24.77 / 18.33 & 14.79 / 8.58 & \textbf{\underline{15.01} / \underline{8.88}} & \textbf{18.36} / 15.65 \\
\bottomrule
\end{tabular}
}
\label{tab:tab4}
\end{table*}

\begin{table*}[htbp]
\centering
\caption{OOD generalization on external datasets using XResNet1d101 (SBP / DBP). Vertical and horizontal subsets represent the training and test sets, respectively. \added{For every setting, we underlined the best-performing model and boldface models that do not perform statistically significantly worse than this model.} }

\scalebox{0.9}{
\renewcommand{\arraystretch}{1.2}
\begin{tabular}{llcccc}
& & \multicolumn{4}{c}{\textbf{Model is tested on}} \\
\cmidrule(lr){3-6}
\multicolumn{2}{l}{\hspace{2.5em}\textbf{\textit{Model is trained on}}} & \textbf{Sensors $\downarrow$} & \textbf{UCI $\downarrow$} & \textbf{PPGBP $\downarrow$} & \textbf{BCG $\downarrow$}  \\
\midrule
\multirow{3}{*}{\textbf{\textit{Combined}}} 
& \textbf{\textit{Calib}} & 19.26 / 11.33 & 21.22 / 12.20 & \textbf{\underline{18.76}} / 9.44 & 13.36 / 8.12 \\
& \textbf{\textit{CalibFree}} & 21.15 / 9.77 & 24.73 / \textbf{10.36} & 25.03 / \textbf{\underline{8.21}} & 14.99 / 7.08 \\
& \textbf{\textit{AAMI}} & 28.67 / 11.40 & 32.47 / 11.78 & 27.36 / 9.71 & 16.91 / 7.43  \\
\midrule
\multirow{3}{*}{\textbf{\textit{Vital}}} 
& \textbf{\textit{Calib}} & 19.58 / 14.54 & 22.42 / 13.33 & \textbf{19.75} / 9.86 & 18.18 / 12.59 \\
& \textbf{\textit{CalibFree}} & 18.45 / \textbf{\underline{8.61}} & 25.05 / 10.83 & \textbf{18.69 / 8.67}& \textbf{\underline{10.05}} / 6.93 \\
& \textbf{\textit{AAMI}} & \textbf{\underline{16.27}} / 10.65 & \textbf{\underline{19.70}/ \underline{10.35}}& 26.82 / 11.67 & 14.33 / 7.66 \\
\midrule
\multirow{3}{*}{\textbf{\textit{MIMIC}}} 
& \textbf{\textit{Calib}} & 32.86 / 23.77 & 43.72 / 28.31 & 33.33 / 15.65 & 26.95 / 12.33  \\
& \textbf{\textit{CalibFree}} & 35.66 / 13.41 & 40.60 / 13.72 & 35.66 / 11.07 & 17.14 / \textbf{\underline{5.90}} \\
& \textbf{\textit{AAMI}} & 40.93 / 15.63 & 44.92 / 16.26 & 35.74 / 10.57 & 21.02 / 6.54 \\
\bottomrule
\end{tabular}
}
\label{tab:tab5}
\end{table*}

\heading{Dependence on training dataset}
For all considered datasets and training scenarios, models trained and tested on the same data (MIMIC, Vital, and Combined) exhibited the \added{relatively} lower errors due to familiarity with the data. However, we stressed that the ID performance is an overly optimistic measure of the model's generalization performance to other datasets. \added{ Furthermore,} comparing the \added{OOD} performance \added{based on different scenarios} between \deleted{CalibFree} MIMIC and \deleted{CalibFree} Vital shows that the MIMIC model generalizes better overall. \deleted{While the ID performance for CalibFree Vital is quite strong, the (OOD) evaluation on MIMIC is poor.}

\heading{Dependence on training scenario}
The results reveal that the lowest MAE for the  Calib subsets is when the training set contains patients from the same dataset, i.e., when corresponding (or combined) Calib training sets are used. This is the expected behavior, as in this case, the models can profit from memorized patient-specific signal patterns observed during training. Surprisingly, Calib models show a reasonable generalization to unseen patients from the CalibFree or AAMI test sets even though they were not trained \replaced{for}{from} this purpose. Furthermore, \deleted{Table \ref{tab:tab4} shows an exceptional performance on the Calib datasets of MIMIC but is significantly heterogeneous; in other words, though performing very strongly, it gave highly variable results, indicative of its performance variability across different scenarios.} AAMI shows the largest overall errors, both in the intra- and inter-data source comparisons; (e.g., for the AAMI Combined, MAE is \replaced{19.38 mmHg (14.04 mmHg)}{13.26mmHg (8.24mmHg)}). This is again the expected behavior since AAMI assesses the generalization to unseen patients (as CalibFree), but at the same time, for a population covering a broad selection of BP values, i.e., with an inherent mismatch in label distribution compared to the training set distribution.

\subsection{OOD Performance evaluation on external datasets}
\label{sec:IV-C}

This section investigates the OOD performance of models trained on different PulseDB subsets and tested on various external datasets, i.e., Sensors, UCI, PPGBP, and BCG. The results are compiled in Table \ref{tab:tab5}.

\heading{Dependence on training dataset}
A first superficial analysis of Table~\ref{tab:tab5} reveals that, except for one BCG result, only models trained on Vital or Combined show good generalization, in the sense of achieving results within the best performance results. This suggests that the Vital subset of PulseDB, which is also part of Combined, seems to be an important component for good generalization. These results stand \replaced{in contrast}{at tension} with the result from the previous section, which seemed to indicate that MIMIC-based models show a better generalization performance on Vital than Vital-based models when tested on MIMIC.

\heading{Dependence on training scenario} The CalibFree \added{ Vital and AAMI Vital} subsets \deleted{particularly "Vital",} demonstrate strong performances with some of the lowest MAE values across metrics\replaced{, achieving top results in four cases for SBP and three cases for DBP}{: 18.46 mmHg (8.6 mmHg) for Sensors and 10.05 mmHg (6.92 mHg) for BCG}. Moreover, the performance of CalibFree subsets is more coherent when tested across different sources, often outperforming Calib models, especially for Vital datasets. This aligns with expectations as \deleted{per training objective} Calib models were not incentivized to generalize to unseen patients but rather to overfit to patient-specific patterns from the training set. \deleted{The count column further highlights their robustness, with "CalibFree Vital" achieving top-three performances 3 times for SBP and 4 times for DBP. Also, AAMI Vital shows strong performance on external datasets, achieving three top-three results for SBP and two for DBP.} Furthermore, AAMI MIMIC shows a poor performance, which should most likely \replaced {stems from}{rather be attributed to} the dataset and not \replaced{from}{to} the training scenario.

\begin{figure}[htbp]  
  \centering
  \includegraphics[width=1\columnwidth]{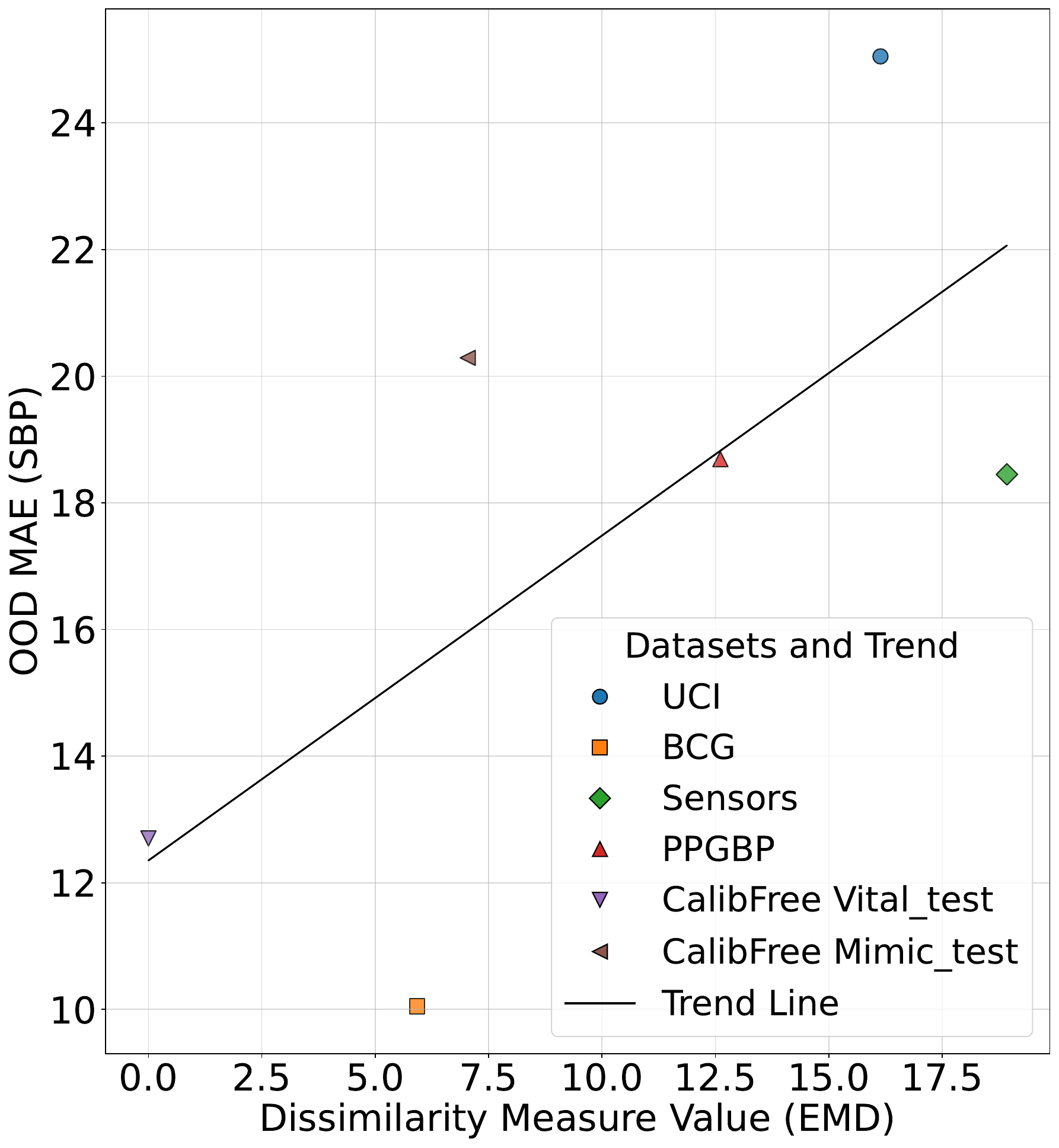}
  \caption{Relationship between Dissimilarity Measure (EMD) and OOD Performance for SBP. The scatter plot shows the correlation between EMD and OOD MAE for SBP. Lower EMD reflects greater similarity to the baseline dataset (CalibFree Vital training set).}
  \label{fig:fig10}
\end{figure}

\heading{Better generalization due to dataset similarity}
At this point, one might hypothesize that the generalization capabilities of models trained on different PulseDB subsets are primarily driven by the similarity between the respective training and evaluation datasets. We investigated this hypothesis for the case of CalibFree Vital as \added{the} training dataset and tested on external datasets that do not share any data with CalibFree Vital, i.e., the four external datasets as well as CalibFree MIMIC. In Figure \ref{fig:fig10}, we presented a scatterplot of the OOD MAE versus dataset similarity quantified via the Earth Mover's Distance (EMD) \cite{pele2009fast} calculated based on the SBP distribution.

Figure \ref{fig:fig10} indeed shows a correlation between dataset dissimilarity (quantified via EMD) and OOD MAE. This suggests that a mismatch between the blood pressure distributions might represent a dominant factor contributing to the domain mismatch between the respective datasets.

\subsection{Sample-Weighting Approach for OOD Generalization}
\label{sec:IV-D}

\heading{Overview} The findings of the previous subsection suggest that the results presented in Table~\ref{tab:tab5} might be biased by dataset similarity, which \replaced{potentially obscuring the identification}{might obfuscate the search} for the training dataset and training scenario that leads to the best generalization. Therefore, this section investigates this hypothesis using the domain adaptation approach introduced in Section~\ref{sec:III_D}.

\begin{table*}[ht]
\centering
\caption{Difference in MAE (weighted - unweighted) for ID and OOD generalization on all categories of PulsedDB dataset using XResNet1d101 (SBP / DBP). Positive values (red) indicate a degradation (increase in MAE), while negative values (green) indicate an improvement (decrease in MAE). The mean improvement through importance weighting across all scenarios is given by 0.36~mmHg for SBP and 0.27~mmHg for DBP.}

\scalebox{0.63}{
\renewcommand{\arraystretch}{1.4}
\begin{tabular}{llccccccccc}

& & \multicolumn{9}{c}{\textbf{Model is tested on}} \\
\cmidrule(lr){3-11}
& & \multicolumn{3}{c}{\textbf{Combined}} & \multicolumn{3}{c}{\textbf{Vital}} & \multicolumn{3}{c}{\textbf{MIMIC}} \\
\cmidrule(lr){3-5} \cmidrule(lr){6-8} \cmidrule(lr){9-11}
\multicolumn{2}{l}{\hspace{2.5em}\textbf{\textit{Model is trained on}}} & \textbf{Calib} & \textbf{CalibFree} & \textbf{AAMI} & \textbf{Calib} & \textbf{CalibFree} & \textbf{AAMI} & \textbf{Calib} & \textbf{CalibFree} & \textbf{AAMI} \\
\midrule
\multirow{3}{*}{\textbf{\textit{Combined}}} 
& \textbf{\textit{Calib}} 
& \diffcellsplitside{nochange}{0.00}{nochange}{0.00}& \diffcellsplitside{degradation}{+0.02}{degradation}{+0.22} 
& \diffcellsplitside{degradation}{+0.45}{improvement}{-0.36} 
& \diffcellsplitside{improvement}{-0.23}{improvement}{-0.12} 
& \diffcellsplitside{improvement}{-0.20}{degradation}{+0.07} 
& \diffcellsplitside{degradation}{+0.34}{improvement}{-0.94} 
& \diffcellsplitside{improvement}{-0.30}{improvement}{-0.01} 
& \diffcellsplitside{degradation}{+0.21}{improvement}{-0.15} 
& \diffcellsplitside{degradation}{+0.54}{degradation}{+0.92} \\
& \textbf{\textit{CalibFree}} 
& \diffcellsplitside{degradation}{+0.55}{degradation}{+0.12} 
& \diffcellsplitside{nochange}{0.00}{nochange}{0.00}& \diffcellsplitside{improvement}{-1.45}{improvement}{-1.62} 
& \diffcellsplitside{improvement}{-0.28}{improvement}{-0.26} 
& \diffcellsplitside{degradation}{+0.56}{improvement}{-0.02} 
& \diffcellsplitside{improvement}{-3.93}{improvement}{-3.65} 
& \diffcellsplitside{improvement}{-0.19}{degradation}{+0.02} 
& \diffcellsplitside{improvement}{-0.10}{improvement}{-0.02} 
& \diffcellsplitside{degradation}{+0.34}{degradation}{+0.74} \\
& \textbf{\textit{AAMI}} 
& \diffcellsplitside{improvement}{-0.19}{degradation}{+0.23} 
& \diffcellsplitside{degradation}{+0.23}{degradation}{+0.32} 
& \diffcellsplitside{nochange}{0.00}{nochange}{0.00}& \diffcellsplitside{improvement}{-3.08}{improvement}{-2.10} 
& \diffcellsplitside{degradation}{+0.28}{degradation}{+0.53} 
& \diffcellsplitside{improvement}{-1.62}{improvement}{-1.12} 
& \diffcellsplitside{improvement}{-0.06}{degradation}{+0.10} 
& \diffcellsplitside{degradation}{+0.26}{degradation}{+0.70} 
& \diffcellsplitside{degradation}{+0.08}{degradation}{+0.70} \\
\midrule
\multirow{3}{*}{\textbf{\textit{Vital}}} 
& \textbf{\textit{Calib}} 
& \diffcellsplitside{degradation}{+0.39}{improvement}{-0.77} 
& \diffcellsplitside{degradation}{+0.32}{improvement}{-0.77} 
& \diffcellsplitside{improvement}{-1.00}{improvement}{-0.60} 
& \diffcellsplitside{nochange}{0.00}{nochange}{0.00}& \diffcellsplitside{improvement}{-0.22}{improvement}{-0.13} 
& \diffcellsplitside{improvement}{-2.47}{improvement}{-1.38} 
& \diffcellsplitside{improvement}{-10.50}{improvement}{-7.18} 
& \diffcellsplitside{improvement}{-0.62}{improvement}{-0.90} 
& \diffcellsplitside{improvement}{-1.11}{degradation}{+0.28} \\
& \textbf{\textit{CalibFree}} 
& \diffcellsplitside{degradation}{+1.30}{degradation}{+0.35} 
& \diffcellsplitside{improvement}{-1.29}{improvement}{-0.43} 
& \diffcellsplitside{improvement}{-0.08}{improvement}{-0.49} 
& \diffcellsplitside{improvement}{-0.39}{degradation}{+0.10} 
& \diffcellsplitside{nochange}{0.00}{nochange}{0.00}& \diffcellsplitside{improvement}{-0.93}{improvement}{-1.24} 
& \diffcellsplitside{improvement}{-0.28}{improvement}{-0.42} 
& \diffcellsplitside{improvement}{-2.12}{improvement}{-0.77} 
& \diffcellsplitside{improvement}{-2.90}{improvement}{-0.49} \\
& \textbf{\textit{AAMI}} 
& \diffcellsplitside{improvement}{-0.13}{improvement}{-0.09} 
& \diffcellsplitside{degradation}{+0.27}{improvement}{-0.03} 
& \diffcellsplitside{degradation}{+0.28}{degradation}{+0.24} 
& \diffcellsplitside{improvement}{-0.28}{improvement}{-0.10} 
& \diffcellsplitside{degradation}{+0.40}{degradation}{+0.29} 
& \diffcellsplitside{nochange}{0.00}{nochange}{0.00}& \diffcellsplitside{degradation}{+0.13}{improvement}{-0.50} 
& \diffcellsplitside{degradation}{+0.43}{improvement}{-0.47} 
& \diffcellsplitside{improvement}{-0.85}{improvement}{-0.41} \\
\midrule
\multirow{3}{*}{\textbf{\textit{MIMIC}}} 
& \textbf{\textit{Calib}} 
& \diffcellsplitside{improvement}{-0.15}{degradation}{+0.27} 
& \diffcellsplitside{degradation}{+0.29}{degradation}{+0.58} 
& \diffcellsplitside{improvement}{-0.98}{improvement}{-0.84} 
& \diffcellsplitside{degradation}{+0.13}{degradation}{+1.04} 
& \diffcellsplitside{improvement}{-0.33}{degradation}{+0.74} 
& \diffcellsplitside{nochange}{0.00}{degradation}{0.09}& \diffcellsplitside{nochange}{0.00}{nochange}{0.00}& \diffcellsplitside{degradation}{+0.22}{improvement}{-0.36} 
& \diffcellsplitside{degradation}{+0.23}{degradation}{+0.64} \\
& \textbf{\textit{CalibFree}} 
& \diffcellsplitside{degradation}{+0.20}{degradation}{+0.16} 
& \diffcellsplitside{improvement}{-0.42}{degradation}{+0.62} 
& \diffcellsplitside{improvement}{-0.38}{improvement}{-1.67} 
& \diffcellsplitside{degradation}{+0.32}{degradation}{+1.18} 
& \diffcellsplitside{improvement}{-0.32}{improvement}{-0.03} 
& \diffcellsplitside{improvement}{-0.63}{improvement}{-4.41} 
& \diffcellsplitside{degradation}{+0.20}{degradation}{+0.01} 
& \diffcellsplitside{nochange}{0.00}{nochange}{0.00}& \diffcellsplitside{improvement}{-0.20}{improvement}{-0.39} \\
& \textbf{\textit{AAMI}} 
& \diffcellsplitside{improvement}{-0.28}{improvement}{+0.50} 
& \diffcellsplitside{degradation}{+1.06}{degradation}{+0.97} 
& \diffcellsplitside{improvement}{-0.56}{improvement}{-0.56} 
& \diffcellsplitside{degradation}{+1.26}{degradation}{+1.29} 
& \diffcellsplitside{degradation}{+1.04}{improvement}{-0.30} 
& \diffcellsplitside{degradation}{+0.73}{degradation}{+0.14} 
& \diffcellsplitside{improvement}{-1.46}{improvement}{-0.66} 
& \diffcellsplitside{degradation}{+0.65}{degradation}{+0.74} 
& \diffcellsplitside{nochange}{0.00}{nochange}{0.00} \\

\bottomrule
\end{tabular}
}
\label{tab:tab_diff_1}
\end{table*}

\heading{Overall impact of importance weighting}
The results of this analysis are compiled in Tables ~\ref{tab:tab_diff_1} and ~\ref{tab:tab_diff_2}, highlighting evaluation results within PulseDB and on external datasets. The tables in the main text \replaced{show}{highlight} the difference between the weighted and unweighted MAE values, whereas the corresponding Tables ~\ref{tab:tab6} and \ref{tab:tab7} in the supplementary material indicate absolute MAE scores achieved via importance weighting.
The results represent direct analogues of Table~\ref{tab:tab4} and Table~\ref{tab:tab5}, however, using sample weights during training. In 81 of 153, i.e., in 53\% of the cases, the importance weighting leads to improved scores. The mean improvement achieved through importance weighting across all scenarios is given by 0.36~mmHg for SBP and 0.27~mmHg for DBP. Under AAMI Vital test sets, our importance weighting approach improved blood pressure estimation accuracy by up to ~4 mmHg. Although modest, these improvements are noteworthy given the strict accuracy requirements of AAMI and represent an important step toward clinical viability. As before, we proceeded by analyzing the results from Table~\ref{tab:tab_diff_1} in more detail.

\heading{Dependence on training dataset}
 When comparing across different datasets, Combined and Vital-based models profit most from importance weighting. Before importance weighting (Table~\ref{tab:tab4}) and after applying it (Table~\ref{tab:tab6}), the best-performing models \replaced{seems to have shifted slightly  from Combined-based models to Vital-based models}{changed: importance weighting seems to slightly shift from Vital-based models to Combined-based models. However, it is worth noting that Combined-based models, when evaluated on Vital or MIMIC are always evaluated ID, as the two are proper subsets of Combined}. Additionally, while importance weighting partially alleviates the aforementioned performance deficiencies of Vital-based models evaluated on MIMIC, it cannot fully \deleted{completely} overcome them.

\heading{Dependence on training and testing scenarios}
In terms of training scenarios, there is no clear trend in the sense of models that \replaced{benefit most}{profit particularly} from importance weighting. However, it is noteworthy\deleted{,} that evaluation on AAMI (with the sole exception of \added{when the model is trained on} AAMI Combined) profits most consistently from importance-weighted training. This aligns with expectations since AAMI is typically furthest from the Calib/Calibfree distributions.

\begin{table*}[ht]
\centering
\caption{Difference in MAE (Weighted - unweighted) for OOD generalization on external datasets (SBP / DBP). Positive values (red) indicate a degradation (increase in MAE), while negative values (green) indicate an improvement (decrease in MAE). The mean improvement through importance weighting across all scenarios is given by 2.66~mmHg for SBP and 0.86~mmHg for DBP.}

\scalebox{0.8}{
\renewcommand{\arraystretch}{1.2}
\begin{tabular}{llcccc} 

& & \multicolumn{4}{c}{\textbf{Model is tested on}} \\ 
\cmidrule(lr){3-6}
\multicolumn{2}{l}{\hspace{2.5em}\textbf{\textit{Model is trained on}}} & \textbf{Sensors} & \textbf{UCI} & \textbf{PPGBP} & \textbf{BCG} \\
\midrule
\multirow{3}{*}{\textbf{\textit{Combined}}} 
& \textbf{\textit{Calib}} 
    & \diffcellsplitside{degradation}{+2.37}{degradation}{+2.88} 
    & \diffcellsplitside{improvement}{-0.39}{degradation}{+0.66} 
    & \diffcellsplitside{degradation}{+1.32}{improvement}{-1.09} 
    & \diffcellsplitside{improvement}{-0.57}{degradation}{+1.03} \\
    
& \textbf{\textit{CalibFree}} 
    & \diffcellsplitside{degradation}{+9.71}{degradation}{+1.73} 
    & \diffcellsplitside{improvement}{-3.86}{degradation}{+3.29} 
    & \diffcellsplitside{degradation}{+7.96}{degradation}{+3.37} 
    & \diffcellsplitside{improvement}{-2.35}{degradation}{+0.27} \\
    
& \textbf{\textit{AAMI}} 
    & \diffcellsplitside{degradation}{+11.25}{degradation}{+0.90} 
    & \diffcellsplitside{degradation}{+13.36}{degradation}{+1.02} 
    & \diffcellsplitside{degradation}{+11.90}{degradation}{+1.69} 
    & \diffcellsplitside{improvement}{-6.27}{improvement}{-0.60} \\
\midrule
\multirow{3}{*}{\textbf{\textit{Vital}}} 
& \textbf{\textit{Calib}} 
    & \diffcellsplitside{improvement}{-1.50}{improvement}{-2.20} 
    & \diffcellsplitside{improvement}{-0.75}{improvement}{-3.03} 
    & \diffcellsplitside{degradation}{+1.73}{degradation}{+1.31} 
    & \diffcellsplitside{improvement}{-6.71}{improvement}{-4.09} \\
    
& \textbf{\textit{CalibFree}} 
    & \diffcellsplitside{degradation}{+1.62}{improvement}{-0.19} 
    & \diffcellsplitside{improvement}{-1.02}{degradation}{+0.25} 
    & \diffcellsplitside{degradation}{+3.06}{degradation}{+1.91} 
    & \diffcellsplitside{degradation}{+0.68}{improvement}{-0.36} \\
    
& \textbf{\textit{AAMI}} 
    & \diffcellsplitside{degradation}{+0.76}{improvement}{-3.08} 
    & \diffcellsplitside{degradation}{+0.06}{improvement}{-1.37} 
    & \diffcellsplitside{improvement}{-9.64}{improvement}{-3.51} 
    & \diffcellsplitside{improvement}{-4.32}{improvement}{-0.15} \\
\midrule
\multirow{3}{*}{\textbf{\textit{MIMIC}}} 
& \textbf{\textit{Calib}} 
    & \diffcellsplitside{improvement}{-13.07}{improvement}{-14.52} 
    & \diffcellsplitside{improvement}{-23.33}{improvement}{-16.87} 
    & \diffcellsplitside{improvement}{-10.60}{improvement}{-4.87} 
    & \diffcellsplitside{improvement}{-11.89}{improvement}{-4.59} \\
    
& \textbf{\textit{CalibFree}} 
    & \diffcellsplitside{improvement}{-12.62}{improvement}{-2.80} 
    & \diffcellsplitside{improvement}{-13.57}{degradation}{+3.32} 
    & \diffcellsplitside{degradation}{+10.09}{degradation}{+6.38} 
    & \diffcellsplitside{improvement}{-7.39}{improvement}{-0.05} \\
    
& \textbf{\textit{AAMI}} 
    & \diffcellsplitside{improvement}{-11.47}{nochange}{0.00} 
    & \diffcellsplitside{improvement}{-22.91}{improvement}{-6.45} 
    & \diffcellsplitside{degradation}{+1.76}{degradation}{+7.46} 
    & \diffcellsplitside{improvement}{-9.19}{degradation}{+1.43} \\

\bottomrule
\end{tabular}
}
\label{tab:tab_diff_2}
\end{table*}

\heading{OOD generalization for models trained with importance weighting}
The impact of the importance weighting approach on OOD generalization on the external datasets is shown in Table \ref{tab:tab_diff_2} and \ref{tab:tab7}. In 40 of the $2\times 9\times 4=72$ cases, i.e., in 55\% of the cases, the approach improved the scores. However, this statistic obfuscates the true picture as it does not take into account the magnitude of the improvement/degradation. The mean improvement through importance weighting across all scenarios is given by 2.66~mmHg for SBP and 0.86~mmHg for DBP, and therefore, substantially larger than the improvements within PulseDB, see Table~\ref{tab:tab_diff_1}. The table shows that the importance of weighting significantly improves the performance of all MIMIC datasets. As before, we proceeded with a more detailed analysis of Table~\ref{tab:tab7}.

\heading{Dependence on training dataset}
These results underscore that choosing the right training data can significantly enhance generalization on specific test sets. For example, training on the Vital dataset achieved 70\% of the best-performing scenarios (highlighted in bold), whereas training on the MIMIC and Combined datasets resulted in only 30\% best-performing models, respectively. Interestingly, the largest and most diverse training dataset, namely Combined, does not \replaced{yield}{lead to} the best generalization.

\heading{Dependence on training scenario}
AAMI subsets, especially ``AAMI Vital'', report lower MAE \replaced{than}{compared to} other subsets, such as 17.03 mmHg (7.57 mmHg) (Sensors), 19.76 mmHg (8.98 mmHg) (UCI), and 17.18 mmHg (8.16 mmHg) (PPGBP), and it acquired one of the best performances on BCG at 10.01 mmHg (7.51 mmHg). This might relate to the broad blood pressure distribution of AAMI, which serves as a basis for importance weighting approaches.

\heading{Comparison to literature results}
A side-by-side comparison between our OOD generalization performance and ID (ML) evaluation results provided by \cite{gonzalez2023benchmark} would provide important insight into the strengths and weaknesses of our approach. Table \ref{tab:tab9} presents the ID results based on ResNet and SpectroResNet \cite{slapnivcar2019blood} models, which are similar architectures to our XResNet1d101 for feature extractions. It is worth noting that the SpectroResNet model integrates a ResNet-GRU architecture that will capture both temporal and spectro-temporal information effectively. It is important to stress that these results report ID performance evaluation and are compared to models trained on PulseDB subsets. To this end, we selected ``AAMI vital'' and ``CalibFree MIMIC'' from tables \ref{tab:tab4},\ref{tab:tab5}, \ref{tab:tab6} and \ref{tab:tab7} as the best-performing models \added{based on} previous analysis. Most notably, models show a solid OOD evaluation performance reaching the performance level of models trained on these dataset\added{s}, i.e., ID performance. This is the case for AAMI Vital, both with and without importance weighting, on Sensors,   AAMI Vital with importance weighting on PPGBP, and CalibFree MIMIC with importance weighting on BCG.

\begin{table*}[t] 
\centering
\caption{Unweighted OOD evaluation vs. importance-weighted OOD evaluation, both trained using XResNetd101 model, in comparison to ID results achieved on the external dataset. \added{The best-performing model for each of the test sets is underlined. As no individual model predictions are available for the ID models, we refrained from an analysis of statistical significance in this case}.}

\scalebox{1}{
\renewcommand{\arraystretch}{1.2}
\begin{tabular}{llcccc}

& & \multicolumn{4}{c}{\textbf{Model is tested on}} \\
\cmidrule(lr){3-6}
\textbf{Experiment Type} & \textbf{Approach} & \textbf{Sensors $\downarrow$} & \textbf{UCI $\downarrow$} & \textbf{PPGBP $\downarrow$} & \textbf{BCG $\downarrow$} \\
\midrule
\multirow{2}{*}{ID (\cite{gonzalez2023benchmark})}
& ResNet & 17.46 / 8.33& \underline{16.59} / \underline{8.30}& 13.62 / 8.61& 12.20 / 7.76 \\
& SpectroResNet & 17.29/ 9.73& 21.92 / 10.21 & \underline{11.01} / 8.46& 9.89 / 6.29\\
\midrule
\multirow{2}{*}{OOD evaluation (Ours)} 

& Trained on AAMI Vital & \underline{16.27} / 10.65& 19.70 / 10.35& 26.82 / 11.67& 14.33 / 7.66\\

& Trained on CalibFree MIMIC & 35.66 / 13.41& 40.60 / 13.72& 35.66 / 11.07& 17.14 / 5.90\\
\midrule
\multirow{2}{*}{Importance weighting (Ours)}

& Trained on AAMI Vital & 17.03 / \underline{7.57}& 19.76 / 8.98& 17.18 / \underline{8.16}& 10.01 / 7.51\\

& Trained on CalibFree MIMIC & 23.04 / 10.61 & 27.03 / 17.04 & 45.75 / 17.45 & \underline{9.75 / 5.85}\\
\bottomrule
\end{tabular}
}
\label{tab:tab9}
\end{table*}

\section{Discussion}
\label{sec:V}
\heading{ID performance does not reflect OOD generalization}
A common trend across all tables is that models achieve consistently lower MAE scores when evaluated on the respective \replaced{corresponding}{matching} test set, i.e., \replaced{under}{when evaluated} ID \added{conditions}. However, this is typically largely exceeded when evaluating on OOD datasets or even in other training scenarios based on the same training dataset. This \replaced{observation clearly highlights}{message nicely } aligns with the findings of \cite{weber2023intensive}, who investigated generalization issues between PPGBP and MIMIC leveraging feature-based models. The magnitude of the performance degradation depends heavily on the training dataset and scenario, but in particular, MIMIC-based models show a poor OOD generalization performance. On the \added{other} hand, there are also combinations of training datasets and scenarios, such as Calibfree Vital or AAMI Vital, that show a particularly good generalization performance. Most \replaced{importantly}{importatnly}, these observations reinforce that ID scores should not be considered as representative of the generalization capabilities of the model to unseen data.

\heading{Effect of domain adaptation}
For a fixed training dataset, the OOD performance showed a correlation with the similarity between training and test datasets assessed via their BP distributions. This \replaced{led}{lead} to the exploration of domain \replaced{adaptation}{adaption} through importance weighting with importance weights inferred from the difference between the two respective BP distributions. The importance weighting approach leads to improved OOD performance.

In general, importance weighting seems to represent an appropriate method to enhance the generalization on unseen external datasets, especially on subsets that represent more variability, such as the CalibFree and AAMI datasets. At this point, it is worth stressing that the presented approach using reweighting based on the BP distribution obviously only captures differences in the BP distributions, whereas the datasets will typically differ according to many other criteria such as patient characteristics, sensor equipment, and/or signal quality.
It is an interesting question for future research if reweighting using a predefined, fixed label distribution, such as a flat or a label distribution inferred from a population cohort, would improve the robustness of the model \replaced{when}{we} encountering an unknown target distribution (without prior knowledge of its label distribution).

\heading{Future research directions}
\added{In this study, we purposely restricted ourselves to the PPG signal as sole model input. This was a deliberate design choice to isolate and evaluate the predictive capacity of the PPG waveform alone, without confounding factors.}
While the presented models showed competitive performance \added{in line with recent benchmark studies \cite{gonzalez2023benchmark,qumphyd1}} and even in comparison to ID performance measures reported in the literature, BP estimation from PPG data \added{alone} remains a challenging task. This can be seen by putting the achieved MAE scores into the perspective of the IEEE standard for cuffless BP estimation \cite{IEEE1708a2019}, where MAEs above 7mmHg are considered as grade D and hence unsuitable for clinical use. In this sense, all presented methods still have a long way to go. However, it is also important to \replaced{note}{stress} that we \deleted{only} \replaced{mostly focused}{reported} \added{on} MAE scores averaged across entire datasets. The absolute error distribution itself is typically \replaced{shows}{a bimodal distribution with} a substantial fraction of samples in the acceptable grades A-B according to the aforementioned IEEE standards and a second group of samples in grade D. Uncovering patterns for the assignment of an unseen sample to one of these groups, for example, based on clinical metadata, would be a large step forward. Next to that, one might rely on the inclusion of additional clinical metadata, more strict data quality control, or pretraining paradigms\added{,} as proposed in the context of self-supervised learning\added{,} to eventually shift the entire MAE distribution in\added{to} a clinically acceptable range. \added{Finally,} it is worth mentioning that the leveraging of pretrained (foundation) models \cite{pillai2024papagei,ding2024siamquality} might represent a promising path to alleviate these issues. The study of the OOD generalization of such models is deferred to future work.

\section{Summary}
\label{sec:VI}

In this study, we conducted a benchmark study on BP prediction from PPG signals using different DL models. In addition, the ID and OOD generalization capability of the DL model for BP prediction is examined across various subsets and data sources. In line with expectations, it can be concluded that the models performed best when trained and tested on the same subset. However, the performance level reached during ID evaluation typically turned out to be an overly optimistic measure of the generalization capabilities when applied to unseen data from unseen sources. In this work, we identified training datasets and scenarios that lead to good OOD generalization. Within PulseDB, it is the Vital subset (in CalibFree and AAMI scenarios) that leads to good generalization performance, whereas MIMIC-based models show poor OOD generalization. This puts into question the use of MIMIC as the predominant training dataset for generalizable BP estimation. 
We identified mismatches in the BP distributions as one important aspect contributing to dataset drifts and explored importance weighting as a domain adaptation technique to mitigate its effect. These techniques establish\added{ed}, in particular, AAMI Vital-based models \replaced{as strong candidates for good}{with good} generalization capabilities.

However, as the most important take-away message, we hope to \replaced{raise awareness in}{sensibilize} the community to the importance of performance evaluation on external datasets and the need for continued work on domain adaptation techniques.

\ifanonymous
\else
\section*{Acknowledgments}
The project (22HLT01 QUMPHY) has received funding from the European Partnership on Metrology, co-financed from the European Union’s Horizon Europe Research and Innovation Programme and by the Participating States. Funding for the University of Cambridge was provided by Innovate UK under the Horizon Europe Guarantee Extension, grant number 10091955. PHC acknowledges funding from the British Heart Foundation (BHF) grant [FS/20/20/34626].

\section*{Data availability}
This work is exclusively based on publicly available datasets, and we released the source code for preprocessing and model training with a citable archived version at Zenodo \cite{moulaeifard2025ppg}

\fi

\bibliographystyle{IEEEtran}
\bibliography{bibfile}

\clearpage
\appendices
\FloatBarrier
\section{Supplementary figures and tables}
\label{sec:supplementary}
\begin{figure*}[htbp]
  \centering

  \includegraphics[width=\textwidth]{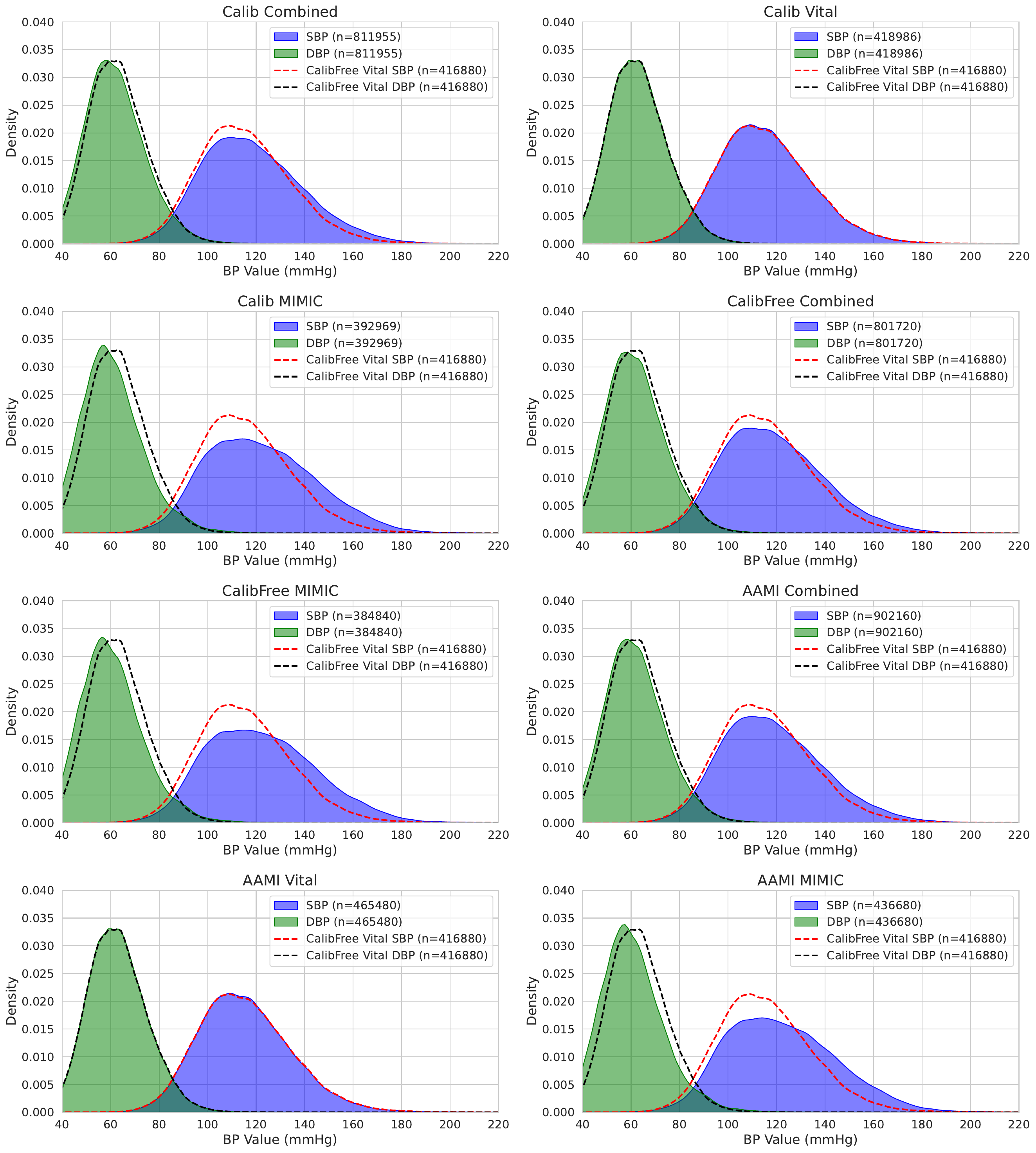}
  \caption{Distribution of SBP and DBP in the \underline{\textbf{training datasets}} across different groups. For reference, the SBP/DBP distribution of the CalibFree Vital \underline{\textbf{training set}} is overlaid in each plot as a dashed red line.}
  \label{fig:Grouped_BP_SBPDBP_Train_RefCalibFreeVital}
\end{figure*}

\begin{figure*}[htbp]
  \centering

  \includegraphics[width=\textwidth]{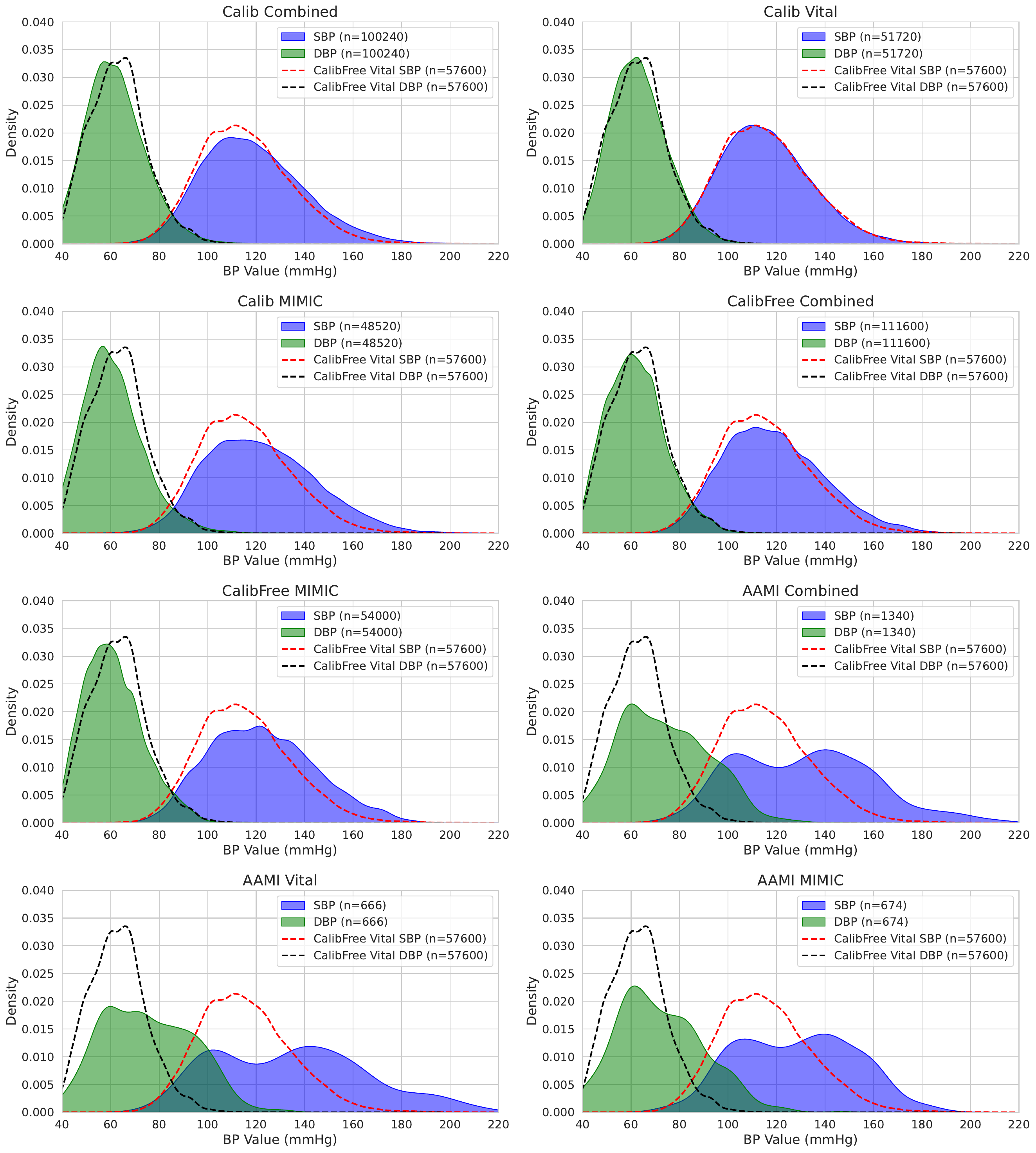}
  \caption{Distribution of SBP and DBP in the \underline{\textbf{test datasets}} across various groupings. The CalibFree Vital \underline{\textbf{test}} distribution is overlaid as a reference (dashed red) to highlight shifts in test distributions.}
  \label{fig:Grouped_BP_SBPDBP_TestOnly_RefCalibFreeVital}
\end{figure*}

\begin{figure*}[htbp]
  \centering

  \includegraphics[width=\textwidth]{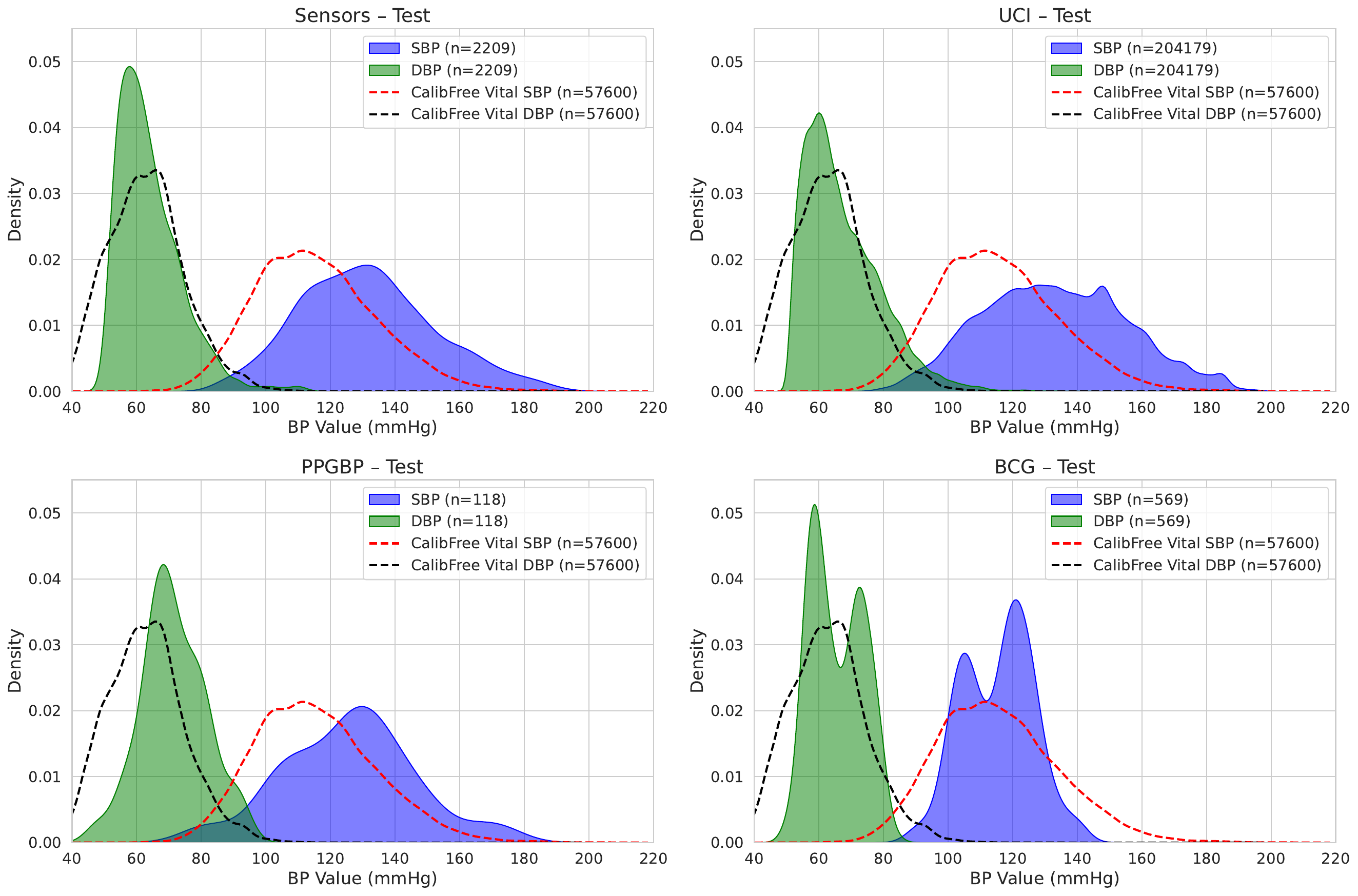}
  \caption{Distribution of SBP and DBP in four \underline{\textbf{external test datasets}}: Sensors, UCI, PPGBP, and BCG. The CalibFree Vital \underline{\textbf{test}} distribution is used as a reference (dashed red) for comparison across datasets.}
  \label{fig:Grouped_BP_SBPDBP_External_RefCalibFreeVital}
\end{figure*}

\begin{figure*}[htbp]
  \centering
  \includegraphics[width=\textwidth]{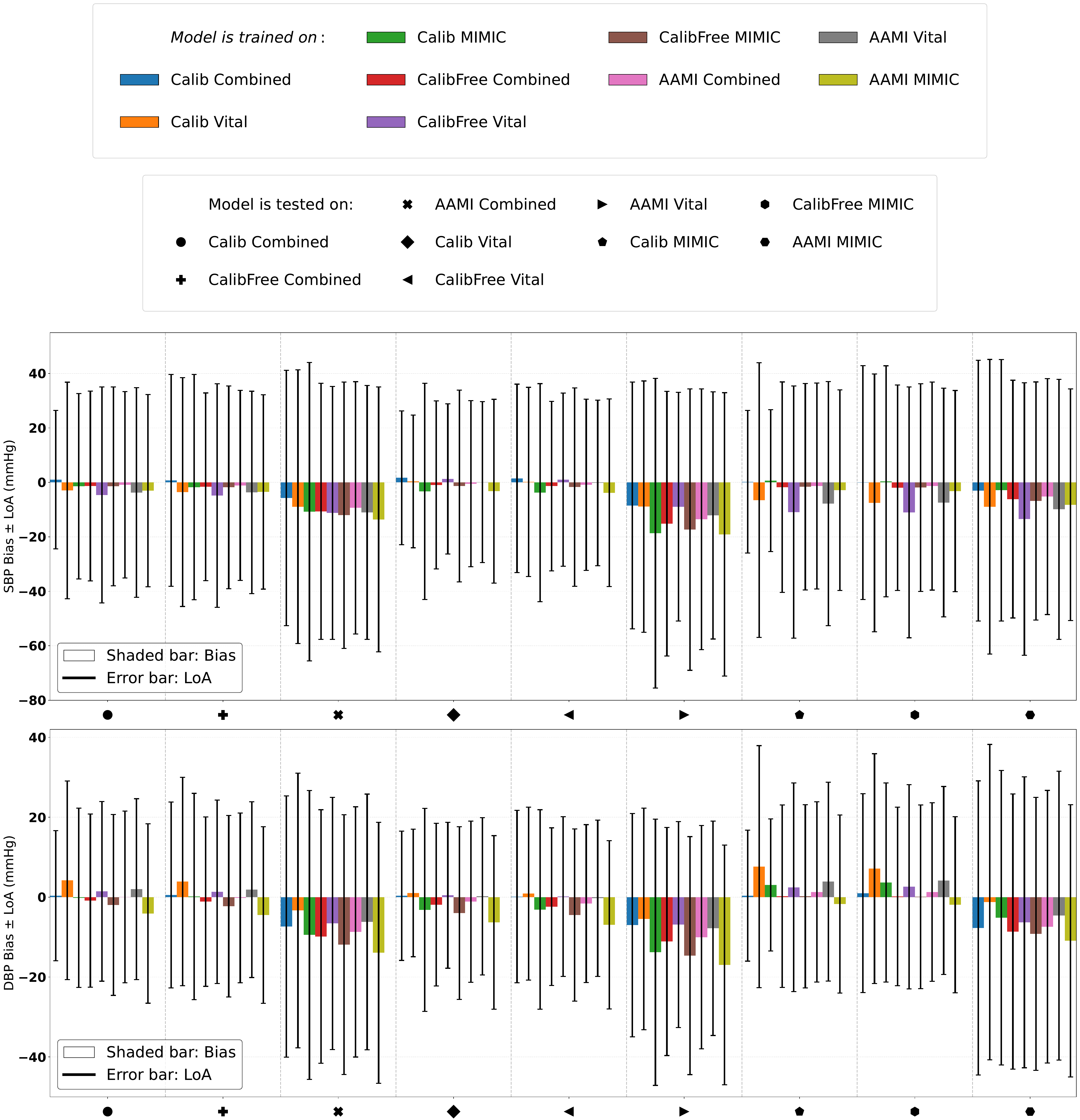}
  \caption{
    Bland-Altman bias with LoA for \underline{\textbf{unweighted}} ID and OOD generalization across all subsets of \underline{\textbf{PulseDB datasets}}. Corresponding MAE values are indicated in Table \ref{tab:tab4}.
  }
  \label{fig:Bias_CI_Internal}
\end{figure*}

\begin{figure*}[htbp]
  \centering
  \includegraphics[width=\textwidth]{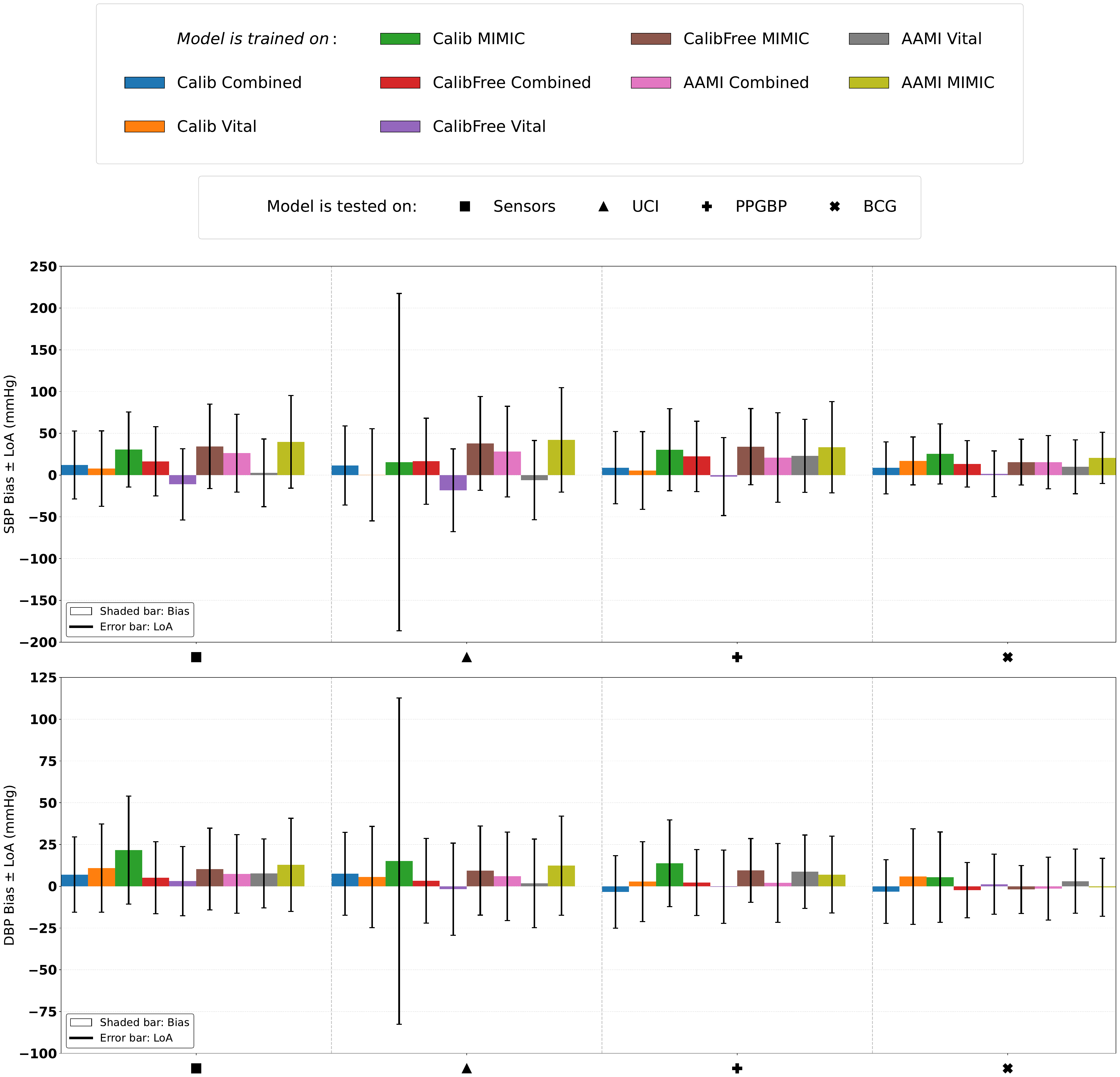}
  \caption{
    Bland-Altman bias with LoA for \underline{\textbf{unweighted}} OOD generalization across all subsets of \underline{\textbf{external datasets}}: BCG, PPGBP, UCI, and Sensors. Corresponding MAE values are indicated in Table \ref{tab:tab5}.
  }
  \label{fig:Bias_CI_External}
\end{figure*}

\begin{figure*}[htbp]
  \centering
  \includegraphics[width=\textwidth]{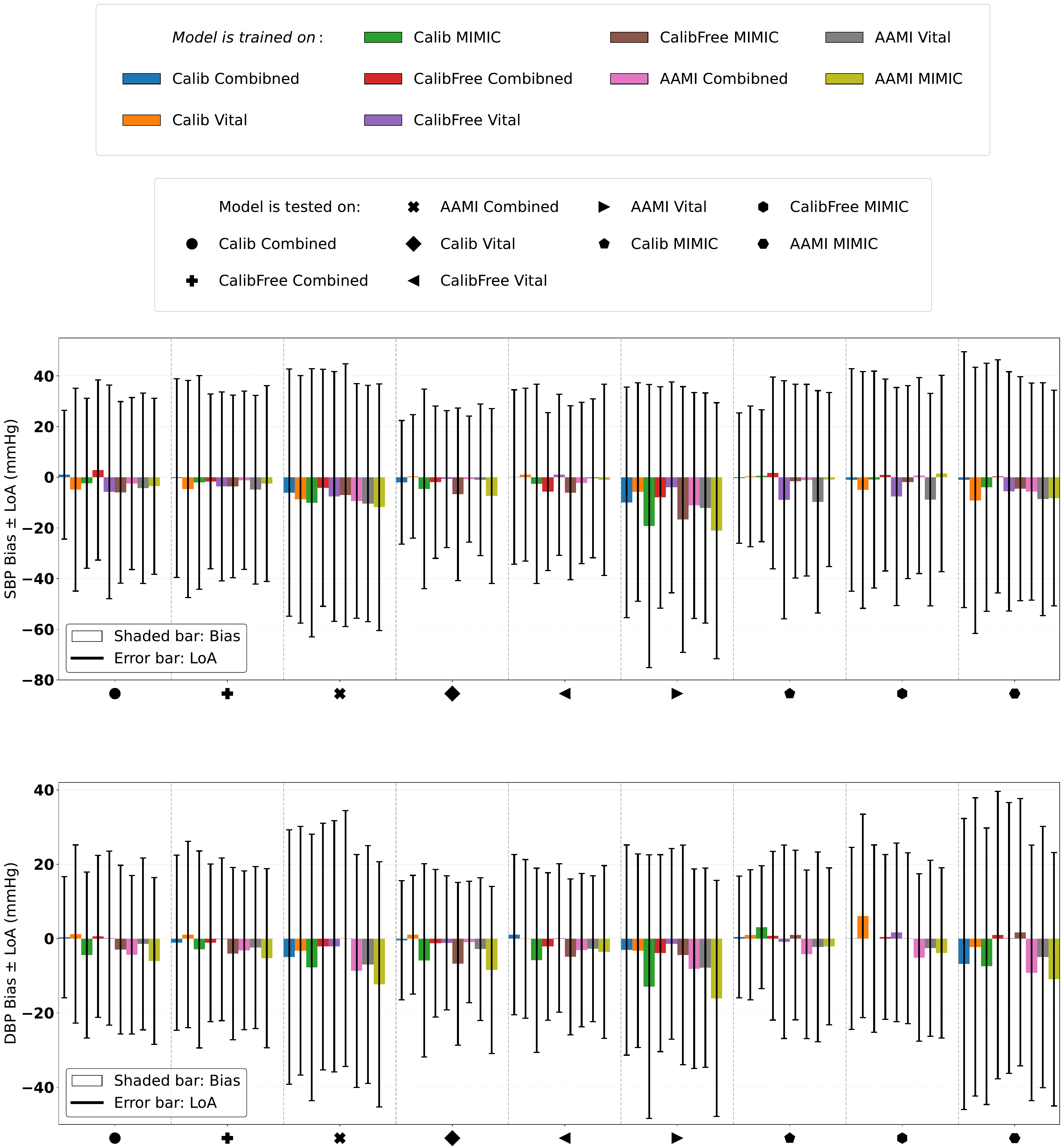}
  \caption{
    Bland-Altman bias with LoA for \underline{\textbf{weighted}} ID and OOD generalization across all subsets of \underline{\textbf{PulseDB datasets}}. Corresponding MAE values are indicated in Table \ref{tab:tab6}. 
  }
  \label{fig:Bias_CI_Internal2}
\end{figure*}

\begin{figure*}[htbp]
  \centering
  \includegraphics[width=1\textwidth]{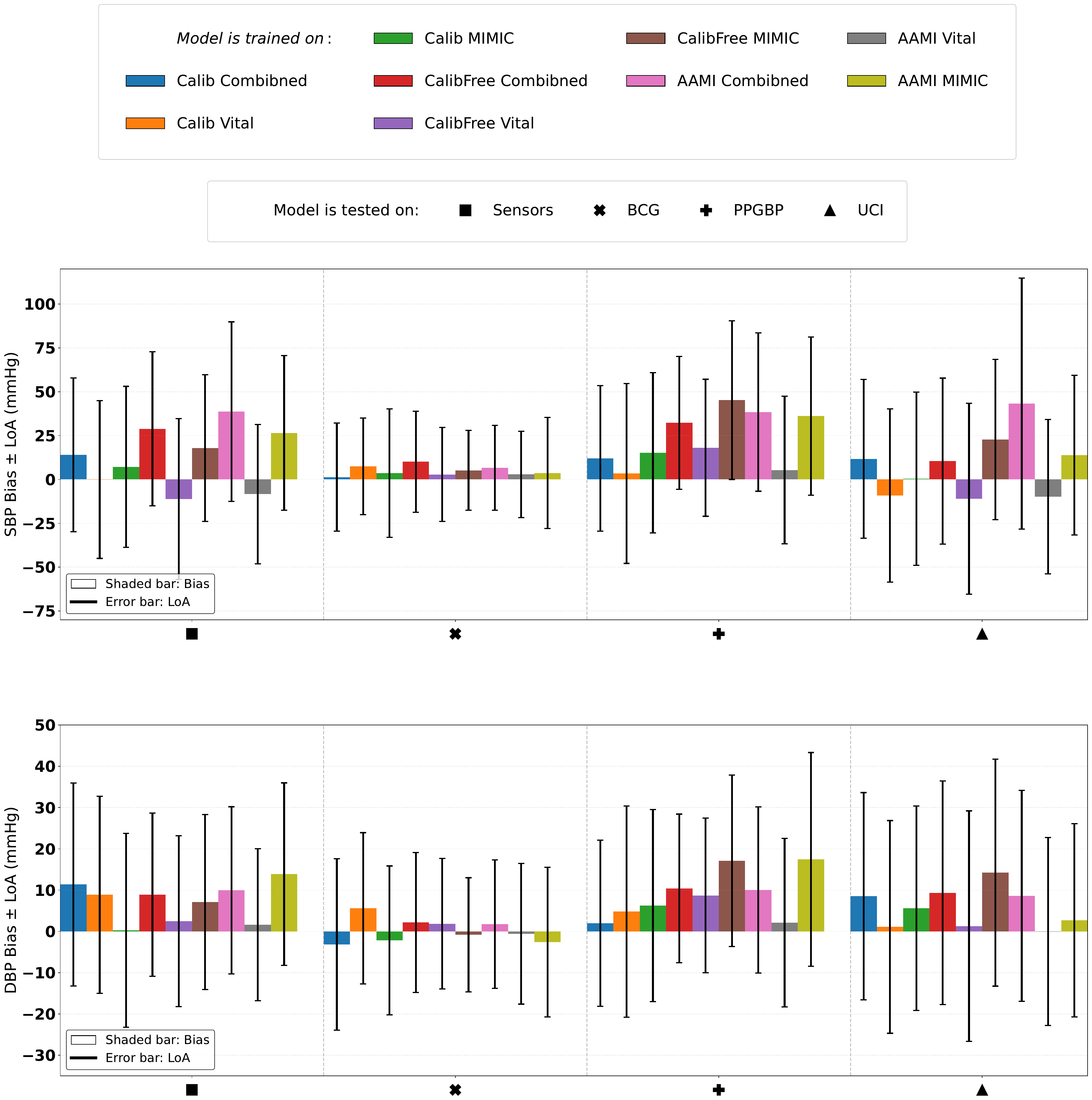}
  \caption{
    Bland-Altman bias with LoA for \underline{\textbf{weighted}} OOD generalization across all subsets of \underline{\textbf{external datasets}}: BCG, PPGBP, UCI, and Sensors. Corresponding MAE values are indicated in Table \ref{tab:tab7}.
  }
  \label{fig:Bias_CI_External2}
\end{figure*}

In Figures 
\ref{fig:Grouped_BP_SBPDBP_Train_RefCalibFreeVital}, \ref{fig:Grouped_BP_SBPDBP_TestOnly_RefCalibFreeVital}, and \ref{fig:Grouped_BP_SBPDBP_External_RefCalibFreeVital}, we visualized the SBP and DBP distributions across the \added{training, internal test, and external test} datasets considered in this work, \added{with the CalibFree Vital training distribution included as a reference in each plot. Also, Tables~\ref{tab:tab6} and \ref{tab:tab7} show absolute performance metrics (as opposed to relative changes in the main text) within PulseDB and on external datasets. Fianlly, Figures \ref{fig:Bias_CI_Internal}, \ref{fig:Bias_CI_External}, \ref{fig:Bias_CI_Internal2}, and \ref{fig:Bias_CI_External2} presents the Bland-Altman analysis results for unweighted and weighted experiments}

\begin{table*}[htbp]
\centering
\caption{Weighted ID and OOD generalization on all categories of PulsedDB dataset using XResNet1d101 (SBP / DBP). For each experiment, boldface denotes the best-performing scenario per subset in terms of the lowest MAE, as well as those not statistically distinguishable from it.}

\scalebox{0.8}{
\renewcommand{\arraystretch}{1.2}
\begin{tabular}{llccccccccc}
 & & \multicolumn{9}{c}{\textbf{Model is tested on}} \\
\cmidrule(lr){3-11}
 & & \multicolumn{3}{c}{\textbf{Combined}} & \multicolumn{3}{c}{\textbf{Vital}} & \multicolumn{3}{c}{\textbf{MIMIC}} \\
\cmidrule(lr){3-5} \cmidrule(lr){6-8} \cmidrule(lr){9-11}
\multicolumn{2}{l}{\hspace{2.5em}\textbf{\textit{Model is trained on}}} & \textbf{Calib $\downarrow$} & \textbf{CalibFree $\downarrow$} & \textbf{AAMI $\downarrow$} & \textbf{Calib $\downarrow$} & \textbf{CalibFree $\downarrow$} & \textbf{AAMI $\downarrow$} & \textbf{Calib $\downarrow$} & \textbf{CalibFree $\downarrow$} & \textbf{AAMI $\downarrow$} \\
\midrule
\multirow{3}{*}{\textbf{\textit{Combined}}} 
& \textbf{\textit{Calib}} & \textbf{9.43 / 5.98}& 15.55 / 9.50 & 19.65 / 13.86& \textbf{9.05 / 5.96}& 13.77 / 8.72 & 18.91 / 11.52 & \textbf{9.29 / 5.85}& 17.41 / 9.80 & 20.36 / 16.89 \\
& \textbf{\textit{CalibFree}} & 14.42 / 8.65& \textbf{13.97 / 8.51}& \textbf{18.65 / 13.11}& 12.06 / 7.97 & 13.14 / \textbf{8.08}& 17.82 / 10.77& 15.32 / 8.87 & \textbf{15.35 / 8.93}& \textbf{18.81} / 15.78\\
& \textbf{\textit{AAMI}} & 13.42 / 8.73& 14.19 / 8.84& \textbf{19.38} / 14.04& \textbf{9.11} / 6.02& 12.84 / 8.55& 18.85 / 12.48 & 15.06 / 9.00 & 15.71 / 9.74 & \textbf{18.38} / 15.17\\
\midrule
\multirow{3}{*}{\textbf{\textit{Vital}}} 
& \textbf{\textit{Calib}} & 15.08 / 8.97 & 17.03 / 10.03 & 20.32 / \textbf{13.35}& 9.09 / 6.09& 13.70 / 8.61 & \textbf{16.70 / 10.61} & 10.17 / 6.45& 19.06 / 12.09 & 22.34 / 16.16 \\
& \textbf{\textit{CalibFree}} & 16.67 / 9.17 & 15.08 / 8.77& 20.13 / \textbf{13.41} & 10.51 / 7.15 & 12.70 / \textbf{8.05}& \textbf{16.53 / 10.31}& 19.85 / 10.08 & 18.17 / 9.66 & 20.03 / \textbf{14.53} \\
& \textbf{\textit{AAMI}} & 14.99 / 9.04 & 15.09 / 8.86 & 19.94 / 13.68 & 11.83 / 7.81 & \textbf{12.58} / 8.16& 19.31 / 12.33 & 18.74 / 9.93 & 18.07 / 9.51& 20.26 / \textbf{14.12}\\
\midrule
\multirow{3}{*}{\textbf{\textit{MIMIC}}} 
& \textbf{\textit{Calib}} & 12.76 / 8.90& 16.89 / 10.92 & 21.96 / 15.43 & 16.21 / 11.53 & 15.91 / 11.01 & 26.04 / 17.13 & 9.52 / 6.64& 17.21 / 10.06 & 20.10 / 16.14 \\
& \textbf{\textit{CalibFree}} & 14.89 / 9.19 & 14.74 / 9.89& 20.98 / 14.19 & 14.56 / 10.30 & 14.55 / 9.27 & 23.42 / 12.27 & 15.36 / 8.94 & 15.47 / 9.27& \textbf{18.51 / 14.66} \\
& \textbf{\textit{AAMI}} & 13.95 / 9.79 & 15.65 / 10.44 & 20.98 / 16.42 & 14.96 / 11.25 & 15.24 / 9.73 & 25.50 / 18.47 & 13.33 / 7.92 & 15.66 / 9.62& \textbf{18.36} / 15.65\\
\bottomrule
\end{tabular}
}
\label{tab:tab6}
\end{table*}

\begin{table*}[htbp]
\centering
\caption{OOD generalization on external datasets for models trained using importance weighting with XResNet1d101 (SBP / DBP). For each subset, boldface marks the scenario achieving the lowest MAE and any others not significantly outperformed by it.}

\scalebox{0.9}{
\renewcommand{\arraystretch}{1.2}
\begin{tabular}{llccccc}
& & \multicolumn{4}{c}{\textbf{Model is tested on}}  \\
\cmidrule(lr){3-6}
\multicolumn{2}{l}{\hspace{2.5em}\textbf{\textit{Model is trained on}}}& \textbf{Sensors $\downarrow$} & \textbf{UCI $\downarrow$} & \textbf{PPGBP $\downarrow$} & \textbf{BCG $\downarrow$}  \\
\midrule
\multirow{3}{*}{\textbf{\textit{Combined}}} 
& \textbf{\textit{Calib}} & 21.63 / 14.21 & 20.83 / 12.86 & 20.08 / \textbf{8.35}& 12.79 / 9.15 \\
& \textbf{\textit{CalibFree}} & 30.86 / 11.50 & 20.87 / 13.65 & 32.99 / 11.58 & 12.64 / 7.35  \\
& \textbf{\textit{AAMI}} & 39.92 / 12.30 & 45.83 / 12.80 & 39.26 / 11.40 & 10.64 / 6.83\\
\midrule
\multirow{3}{*}{\textbf{\textit{Vital}}} 
& \textbf{\textit{Calib}} & 18.08 / 12.34& 21.67 / 10.30& 21.48 / 11.17 & 11.47 / 8.50 \\
& \textbf{\textit{CalibFree}} & 20.07 / 8.42& 24.03 / 11.08 & 21.75 / 10.58& 10.73 / 6.57\\
& \textbf{\textit{AAMI}} & \textbf{17.03 / 7.57}& \textbf{19.76 / 8.98}& \textbf{17.18 / 8.16}& \textbf{10.01} / 7.51 \\
\midrule
\multirow{3}{*}{\textbf{\textit{MIMIC}}} 
& \textbf{\textit{Calib}} & 19.79 / 9.25& 20.39 / 11.44& 22.73 / 10.78 & 15.06 / 7.74 \\
& \textbf{\textit{CalibFree}} & 23.04 / 10.61 & 27.03 / 17.04 & 45.75 / 17.45 & \textbf{9.75 / 5.85} \\
& \textbf{\textit{AAMI}} & 29.46 / 15.63 & 22.01 / 9.81& 37.50 / 18.03 & 11.83 / 7.97 \\
\bottomrule
\end{tabular}
}
\label{tab:tab7}
\end{table*}

\end{document}